\pdfoutput=1
\documentclass[9.5pt,journal,compsoc,cspaper]{IEEEtran}
\usepackage{amsmath,amsfonts}

\usepackage{ragged2e}
\usepackage{algorithmic}
\usepackage{algorithm}
\usepackage{array}
\usepackage[caption=false,font=normalsize,labelfont=sf,textfont=sf]{subfig}
\usepackage{textcomp}
\usepackage{stfloats}
\usepackage{url}
\usepackage{verbatim, booktabs}
\usepackage{graphicx, multirow}
\usepackage{wrapfig}
\usepackage{threeparttable}
\usepackage{siunitx}    
\usepackage{cite}
\usepackage[breaklinks,colorlinks,pagebackref]{hyperref}
\usepackage{color, xspace}

\makeatletter
\DeclareRobustCommand\onedot{\futurelet\@let@token\@onedot}
\def\@onedot{\ifx\@let@token.\else.\null\fi\xspace}
\def\eg{\emph{e.g}\onedot} 
\def\ie{\emph{i.e}\onedot} 
 
\def\etc{\emph{etc}\onedot}

\makeatother

\begin{document}
\title{DragOSM: Extract Building Roofs and Footprints from Aerial Images by Aligning Historical Labels}
\author{Kai Li, Xingxing Weng, Yupeng Deng, Yu Meng, Chao Pang, Gui-Song Xia, Xiangyu Zhao

\IEEEcompsocitemizethanks{\IEEEcompsocthanksitem K. Li is with the School of Electronic, Electrical and Communication Engineering, University of Chinese Academy of Sciences, China, and also with the Department of Data Science, City University of Hong Kong, Hong Kong. E-mail: likai211@mails.ucas.ac.cn.
\IEEEcompsocthanksitem X. Weng, C. Pang, and G. S. Xia are with the School of Artificial Intelligence, Wuhan University, China.
E-mail: \{pangchao, xingxingw, guisong.xia\}@whu.edu.cn.
\IEEEcompsocthanksitem Y. Deng, Y. Meng are with the Aerospace Information Research Institute, Chinese Academy of Sciences, China. E-mail: \{dengyp, mengyu\}@aircas.ac.cn. 
\IEEEcompsocthanksitem X. Zhao is with the Department of Data Science, City University of Hong Kong, Hong Kong. E-mail: xianzhao@cityu.edu.hk. 
\IEEEcompsocthanksitem The studies in this paper have been supported by the Strategic Priority Research Program of the Chinese Academy of Sciences, Grant No.XDA0360303.
\IEEEcompsocthanksitem Corresponding authors: Y. Deng, Y. Meng.
}}

\markboth{Journal of \LaTeX\ Class Files, Preprint, Feb~2025}%
{Shell \MakeLowercase{\textit{et al.}}: A Sample Article Using IEEEtran.cls for IEEE Journals}


\IEEEtitleabstractindextext{
\begin{abstract}
\justifying
Extracting polygonal roofs and footprints from remote sensing images is critical  for large-scale urban analysis. 
Most existing methods rely on segmentation-based models that assume clear semantic boundaries of roofs, but these approaches struggle in off-nadir images, where the roof and footprint are significantly displaced, and facade pixels are fused with the roof boundary. 
With the increasing availability of open vector map annotations, \eg, OpenStreetMap, utilizing historical labels for off-nadir image annotation has become viable because remote sensing images are georeferenced once captured. However, these historical labels commonly suffer from significant positional discrepancies with new images and only have one annotation (roof or footprint), which fails to describe the correct structures of a building. 
To address these discrepancies, we first introduce a concept of an alignment token, which encodes the correction vector to guide the label correction. Based on this concept, we then propose Drag OpenStreetMap Labels (DragOSM), a novel model designed to align dislocated historical labels with roofs and footprints. 
Specifically, DragOSM formulates the label alignment as an interactive denoising process, modeling the positional discrepancy as a Gaussian distribution. During training, it learns to correct these errors by simulating misalignment with random Gaussian perturbations; during inference, it iteratively refines the positions of input labels.
To validate our method, we further present a new dataset, Repairing Buildings in OSM (ReBO), comprising 179,265 buildings with both OpenStreetMap and manually corrected annotations across 5,473 images from 41 cities. Experimental results on ReBO demonstrate the effectiveness of DragOSM. Code, dataset, and trained models are publicly available at~\url{https://github.com/likaiucas/DragOSM.git}.


\end{abstract}

\begin{IEEEkeywords}
Building footprint extraction, building roof extraction, learning offset vector, off-nadir aerial image. 
\end{IEEEkeywords}

}

\maketitle

\bstctlcite{BSTcontrol}

\section{Introduction}

\begin{figure}
    \centering
    \includegraphics[width=0.9\linewidth]{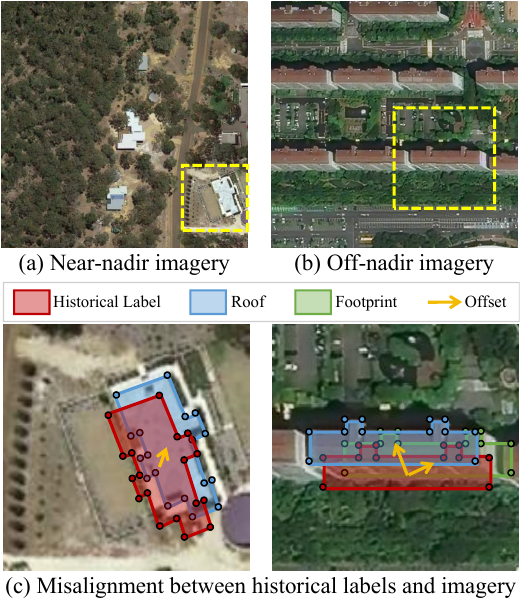}
    \caption{ Comparison of the label alignment problem. (a) In near-nadir imagery, the roof and footprint largely overlap. (b) In off-nadir imagery, the roof and footprint are displaced. (c) Historical labels become misaligned with updated imagery due to georeferencing shifts and human factors. This misalignment is more complex in off-nadir views (b), as the roof-footprint displacement prevents realignment with a single, uniform correction, unlike in near-nadir views (a). 
    Especially for tall buildings, the roof and the historical label can become completely disjoint, making it difficult to establish the correct correspondence.
    }\label{fig:intro}
\end{figure}

\IEEEPARstart{E}{xtracting} footprints and roofs of buildings from remote sensing imagery as vector polygons is a fundamental task in geoinformatics~\cite{bfe_review}.
Preliminary studies on building polygon extraction primarily focus on near-nadir imagery~\cite{structureapproch, geofeature, vit_building_extraction, hisup, sampolybuild}, where the roofs and footprints of buildings are well aligned (see Fig.\ref{fig:intro}(a)). 
This characteristic allows these methods to simplify the polygon extraction by targeting either the roof or the footprint, typically the roof, owing to its visibility. The roof polygon extraction is commonly performed in two stages: first, semantic segmentation is applied to produce raster masks, followed by the vectorization of these masks into polygons~\cite{structureapproch, geofeature, vit_building_extraction}. Regarding the issue of the gap between predicted masks and polygons in two-step methods, recent works like HiSup~\cite{hisup} and SAMPolyBuild~\cite{sampolybuild} shift from mask-to-polygon conversion to mask-guided connection of boundary keypoints, thereby acquiring building polygons with higher geometric fidelity. 

Despite considerable progress in building polygon extraction, state-of-the-art methods are predominantly effective only on near-nadir imagery and face significant limitations when applied to off-nadir imagery~\cite{BONAI}. As shown in Fig.\ref{fig:intro}(b), off-nadir imagery exhibits noticeable displacement between building roofs and their corresponding footprints, violating the alignment assumption inherent in most existing methods. This displacement forces current methods to process roofs and footprints separately. During roof extraction, semantic confusion between roof and facade pixels often leads to inaccurate roof outlines. 
During footprint inference, occlusion from building facades results in incomplete footprint masks or missing boundary keypoints. This situation becomes particularly challenging in dense urban blocks, especially when buildings cast significant shadows. Moreover, the independent processing of roofs and footprints hinders the establishment of explicit structural correspondences between them, reducing the practicality of these methods in real-world applications (\eg, 3D city modeling).  

To handle off-nadir imagery, specialized methods such as LOFT~\cite{BONAI},  MLS-BRN~\cite{MLS-BRN}, OBM~\cite{obm}, and PolyFootNet~\cite{li2024polyfootnet} have been developed. They leverage the geometric similarity between roof and footprint polygons by simultaneously predicting the roof polygon and its offset relative to the corresponding footprint. The estimated offset is subsequently used to locate the footprint polygon. While these methods demonstrate improved performance compared to earlier near-nadir-focused approaches, they remain susceptible to semantic confusion between roof and facade elements, limiting their performance in building polygon extraction. Additionally, their reliance on post-processing algorithms, which are either sensitive to parameter selection~\cite{BONAI, MLS-BRN} or demand tedious manual intervention~\cite{obm}, compromises their practical utility in large-scale scenarios.

Therefore, we consider leveraging historical labels, which are commonly human-annotated with clear boundary information, to circumvent the inherent issues of semantic segmentation-based approaches. Coincidentally, a fundamental characteristic of remote sensing imagery is its inherent metadata, \eg, acquisition time and geolocation. Using this information, the corresponding area can be easily located on open-source, continuously updated global maps (\emph{e.g.}, OpenStreetMap\footnote{https://www.openstreetmap.org/} and Google Maps\footnote {https://www.google.com/maps/}), where ground objects are represented with vector primitives, namely points, lines, and polygons. This observation motivates us to leverage this pre-vectorized object information, referred to as historical labels in this paper, to facilitate building polygon extraction.

Although historical labels are carefully retrieved via geolocation, they remain poorly aligned with the remote sensing imagery. As shown in Fig.\ref{fig:intro}(b) and (c), significant positional discrepancies are observed between these labels and the actual building polygons. This misalignment is especially pronounced in off-nadir imagery. Two primary factors contribute to this issue: (1) errors introduced during the georeferencing process of remote sensing imagery, such as inaccurate selection of ground control points and the use of inappropriate geometric transformation models~\cite{snyder1987mapprojection, zitova2003imageregistration, le2011imageregistration}; (2) historical labels are updated via a crowdsourced annotation mode, a process whose quality is influenced by human subjectivity and inherent errors in the source data, potentially leading to inaccuracies.


Building upon this insight, we reformulate building polygon extraction as an alignment problem between historical labels and remote sensing imagery. State-of-the-art methods address this alignment by predicting a global flow field to rectify keypoints in historical labels~\cite{Zampieri_2018_ECCV, Girard_2018_ACCV, chen2019autocorrect, Girard_2019_IGARSS}. In near-nadir imagery, the substantial overlap between building footprints and roofs, combined with the slight positional discrepancies between historical labels and actual buildings, allows these methods to yield promising results through a single correction step. However, they often encounter significant performance degradation when tasked with off-nadir imagery. As seen in Fig.\ref{fig:intro}(c), variations in building heights result in diverse and potentially large positional discrepancies between historical labels and roof polygons. Moreover, a separate discrepancy exists between labels and footprints, which often differs in both direction and magnitude from the label-to-roof discrepancy.

In this paper, we devote ourselves to developing an effective alignment framework capable of handling both near-nadir and off-nadir imagery to achieve accurate building polygon extraction. To this end, we propose DragOSM, which aligns historical labels with building footprints and roofs in remote sensing imagery via a novel ``dragging'' mechanism. This operation is enabled by \textit{alignment tokens}, a representation we introduce to encode positional discrepancies. Decoding these tokens produces displacement offsets that guide the dragging process. While separate alignment tokens could intuitively be defined for label-to-roof and label-to-footprint dragging, we consider that historical labels tend to be annotated closer to footprints. Accordingly, we design two tokens: one for label-to-footprint alignment, which drags labels to footprint polygons, and another for footprint-to-roof alignment, which further shifts these into roof polygons.



\begin{figure}
    \centering
    \includegraphics[width=1\linewidth]{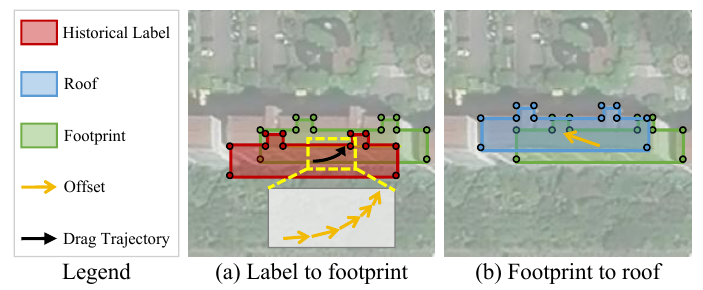}
    \caption{
    The figure illustrates the proposed two-stage inference process. First (a), a continuous denoising process progressively drags a noisy historical label to align with the building's footprint. Second (b), the corrected footprint is translated to the roof's location in one step.
    }
    \label{fig:visual_train_test}
\end{figure}

To instantiate the dragging mechanism, we model the positional discrepancy as a Gaussian distribution. During each training iteration, we simulate diverse misalignment cases between historical labels and imagery by perturbing ground-truth footprint polygons with random Gaussian noise. Compared to directly using real labels (obtained from open-source maps), this strategy enhances DragOSM's adaptability to misalignments from varying imaging viewpoints and building heights. Given the input imagery and noisy label, DragOSM enables alignment tokens to interact with both, encoding positional discrepancies. These tokens are then decoded into displacement offsets, forming a denoising process. Since real-world misalignments are often unpredictable and potentially extreme, single-step denoising is insufficient. Fortunately, DragOSM's architecture natively supports iterative denoising, in which the label corrected by the offset from the current step becomes the input for the next denoising step. By accumulating the offsets generated in each denoising step, a trajectory is formed that progressively shifts the historical label toward the building footprint, as visualized in Fig.\ref{fig:visual_train_test}. This capability enables DragOSM to effectively handle the significant positional discrepancies commonly found in off-nadir imagery.




Additionally, we conducted a detailed analysis of the convergence behavior in iterative denoising. Our observations indicate that the number of iterations required for denoising to converge to a local optimum is related to the predefined accumulation schedules. This schedule governs the proportion of the $t$-th predicted offset that is incorporated at each step. Once this schedule is set, convergence steps can be estimated. Furthermore, we introduce two simple yet effective test-time strategies applied during the denoising iterations. The first involves adding perturbations after a fixed number of denoising steps to generate multiple local optima through repeated denoising cycles, while the second continues denoising beyond an estimated convergence step to collect additional local optima from subsequent steps. In both strategies, the collected local optima are averaged to produce a robust and accurate correction. Leveraging these augmentations, we develop two advanced versions of our framework: DragOSM-t1, which incorporates noise perturbations, and DragOSM-t1.5, which employs convergence-aware denoising.

To train and evaluate our DragOSM, we create a new building polygon extraction dataset called \textbf{ReBO} (Repairing Building in OpenStreetMap). 
Each sample in ReBO consists of remote sensing imagery paired with a corresponding historical label from OpenStreetMap, along with meticulously annotated ground-truth polygons for both roofs and footprints. The dataset contains 5,473 images and 179,265 building instances, spanning 41 cities across 22 countries on 6 continents.   

The main contributions of this paper are summarized as follows:
\begin{itemize}

    \item We reformulate polygonal building extraction as a task of aligning historical labels with the remote sensing imagery, thereby enabling a unified method that performs effectively on both near-nadir and off-nadir imagery.
    \item With the new formulation, we present \textit{DragOSM}, a novel framework that aligns labels with imagery via a dragging mechanism. This process is driven by \textit{alignment tokens}, a new concept we introduce to encode positional discrepancies by interacting with both input labels and imagery.
    \item To improve the adaptability of DragOSM to diverse misalignments, we model positional discrepancies as a Gaussian distribution. During training, random Gaussian perturbations simulate misalignments to guide the model denoising. During inference, this denoising function is applied iteratively, which further enhances the accuracy of polygon extraction.
    \item We introduce \textit{ReBO}, a building polygon extraction dataset comprising remote sensing imagery paired with both historical and well-aligned labels. This dataset facilitates the advancement of extraction methods based on the alignment idea. 

\end{itemize}
\section{Related Work}
This section reviews prior work from two perspectives. First, as our paper's core task is to provide vectorized building labels, Sec.\ref{sec:bfe} details the evolution of roof and footprint extraction methods. Second, given that our method is based on a denoising framework, Sec.\ref{dntraining} compares the specific denoising design of DragOSM against those employed in other related tasks.
\subsection{Extracting Roofs and Footprints}
\label{sec:bfe}
The simultaneous extraction of building roof-footprint pairs from remote sensing imagery is a task of significant importance. Previously, research in this area has primarily focused on extracting roofs from near-nadir images~\cite{li2023joint, hisup, sampolybuild}. In such imagery, the vertical viewing angle causes the roof and footprint to be largely overlapped, allowing the extracted roof to be used as a proxy for the footprint. 

Advances in imaging technology have led to increasingly clear remote sensing images of buildings. However, this progress is often accompanied by a greater sensitivity to satellite viewing angles, as oblique (off-nadir) perspectives that capture both the roof and facade become more common. While these views provide richer textural details, \eg, building external features, height information, \etc, they also introduce a significant challenge: the semantic boundary between the roof and the facade becomes ambiguous. This ambiguity complicates the simultaneous extraction of well-defined roofs and their corresponding footprints. 

To address the challenges of view-angle displacement in off-nadir imagery, several methods have been proposed to explicitly model the relationship between roofs and footprints. MTBRNet~\cite{MTBRNet} pioneered a classification-based approach by encoding spatial displacements as one-hot vectors. Building on this, LOFT~\cite{BONAI} instantiated the offset concept as a direct regression target, combining an instance segmentation network with a vector prediction head to create a more intuitive model. To address the scarcity of datasets with explicit displacement labels, MLS-BRN~\cite{MLS-BRN} further extended the LOFT framework by incorporating an auxiliary height prediction task, thereby enhancing its ability to predict both roofs and footprints.

Another line of work has sought to bypass the difficulties of roof extraction altogether. To resolve the ambiguous semantic boundary between roofs and facades, PolyFootNet~\cite{li2024polyfootnet} explored a \textit{Building Segmentation + Offset} strategy. Instead of segmenting the roof directly, these methods segment the entire, more easily identifiable building boundary and then predict an offset to derive the roof and footprint. While this approach cleverly circumvents the roof-facade ambiguity, the stochastic nature of the mask prediction process can lead to a lack of morphological and spatio-temporal consistency in the final outputs.

Departing from previous approaches, we propose a method that first retrieves historical labels using the inherent geographic information of remote sensing imagery, and then obtains image-aligned roof and footprint annotations through continuous positional correction. Our model, DragOSM, unifies all positional offsets under the single concept of an \textit{alignment token}.
This methodology provides two distinct advantages: (1) \textbf{High-Quality Priors: }By leveraging historical labels, which are typically manually annotated, our method starts with a high-quality initial guess that already possesses clean semantic boundaries and reliable contour information. (2) \textbf{Architectural Consistency:} DragOSM is designed as a label-in-label-out model where the input and output share the same modality. This iterative architecture ensures exceptional consistency and robustness throughout the refinement process.

\subsection{Noise Training}
\label{dntraining}


Noise augmentation has become a common strategy in the training of detection transformer (DETR)~\cite{detr}-based models, as it helps improve convergence and model robustness.
\eg, models in the Mask DINO~\cite{dn-detr, dino, maskdino} family commonly employ label flipping strategies, such as directly switching the type of a correct class token, during training to introduce richer forms of noise. This approach accelerates the decoder’s understanding and convergence with respect to various types of abstract label tokens.

On the other hand, noise injection and denoising training strategies are also widely employed in the training of generative AI (GenAI) models. The core idea is to incrementally train and infer starting from pure noise, ultimately generating images~\cite{NEURIPS2020_ddpm}, videos~\cite{ma2025sayanything}, text~\cite{nie2025lldm}, or multimodal context~\cite{xie2024showo} that conform to meaningful data distributions.

In the label alignment problem, the training process of MapAlignment~\cite{Zampieri_2018_ECCV, Girard_2018_ACCV, Girard_2019_IGARSS} involves adding a uniform Gaussian noise to the raster ground-truth labels before training to create a fixed target vector field. This use of static noise, however, struggles to realistically simulate the complex nature of real-world positional errors. Furthermore, this approach is fundamentally incompatible with off-nadir scenarios, where the correction field must be non-uniform to handle the height-dependent displacement of each individual building.

Unlike existing methods, DragOSM instantiates the proposed core concept of the \textit{alignment token} and interprets the positional perturbation of OSM labels as a Gaussian process centered at the ground truth. During multi-step inference, the correction process is modeled as the cumulative effect of Gaussian process differentials. This label-in-label-out framework enables DragOSM to explicitly leverage the spatial relationship between labels and imagery, setting it apart from prior noise-based learning methods.

\section{Methodology}
After a general setup of the image-label alignment problem by Sec.\ref{setup}, Sec.\ref{overview} will present a physical structure of our method. Subsequently, Sec.\ref{train} provides a detailed account of our training process, explaining how the model is guided to recognize the Gaussian characteristics of noisy label perturbations and to accurately map these noisy labels to the corresponding roof and footprint locations.
Finally, Sec.\ref{test} describes how we use the DragOSM model to perform inference, mapping noisy labels to accurate roof and footprint annotations.


\begin{figure*}
    \centering
    \includegraphics[width=1\linewidth]{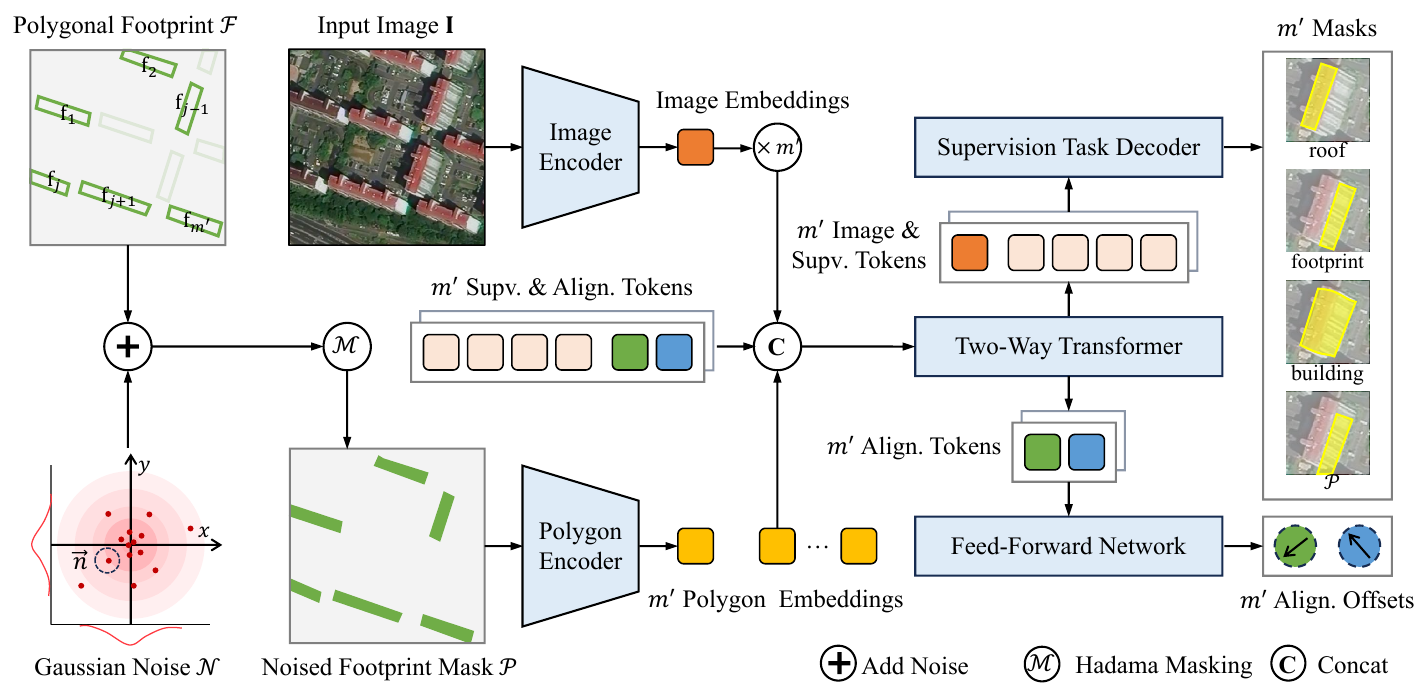}
    \caption{ Overview of the DragOSM Training Process.
During training, ground-truth polygons are perturbed with additive Gaussian noise to simulate noisy historical labels. The model learns to recover the ground-truth positions by decoding \textit{alignment tokens} into corrective offsets based on the image context. Additionally, an auxiliary mask supervision task is used to enhance the model's understanding of building structures. }
    \label{fig:dragosm}
\end{figure*}

\subsection{Problem Setup}
\label{setup}

Given an aerial image $\mathbf{I}$ and its corresponding historical annotations ${\mathcal{P}}$, the image-label alignment problem aims to accurately match and align the polygonal building roof ${\mathcal{R}}$ and footprint ${\mathcal{F}}$ in the image $\mathbf{I}$ with the given ${\mathcal{P}}$. 

In this work, we propose to solve this problem by training an image-label interactive model on given data ${\mathcal{D}}$, \ie,
\begin{equation}
   \mathcal{D}=\left\{\mathbf{I}, \mathcal{P}, \mathcal{R}, \mathcal{F}\right\},
\end{equation}
where $\mathcal{P}, \mathcal{R}, \mathcal{F}$ are the corresponding polygons of buildings contained by the image $\mathbf{I}$. In particular, for each $\mathcal{P}, \mathcal{R}, \mathcal{F}$ consists of $m$ building annotations, including ground truth polygonal roof $\mathcal{R}=\{{\mathbf{r}_j}\}_{j=1}^M$, footprint $\mathcal{F}=\{{\mathbf{f}_j}\}_{j=1}^M$, and OSM polygon $\mathcal{P}=\{{\mathbf{p}_j}\}_{j=1}^M$. 
For each $\mathbf{r}_j$, $\mathbf{f}_j$ and $\mathbf{p}_j$ include a set of keypoints. 

Moreover, the alignment relationships between each two of $\mathcal{R}$, $\mathcal{F}$ and $\mathcal{P}$ are recorded as offsets under Eulerian coordinates, \ie, OSM position to footprint position $\vec{F}=\{\vec{f_j}\}_{j=1}^M$, OSM position to roof position $\vec{R}=\{\vec{r_j}\}_{j=1}^M$, and footprint to roof position $\vec{O}=\{\vec{o_j}\}_{j=1}^M$. 
$\vec{f}_j$, $\vec{r}_j$, and $\vec{o}_j$ are a vector in two dimensions.
Based on the geometric relationship among the three elements, we have: 
\begin{equation}
\label{eq:relationship}
    \vec{f_j} + \vec {o_j} = \vec {r_j}. 
\end{equation}

Eq.\ref{eq:relationship} performs as the bridge between OSM position and ground truth location; in DragOSM, two of three are used to align labels. 
By design, we introduce $\vec{f_j}$ and $\vec {o_j}$ as latent variables responsible for representing the corrective alignment for each label.
Therefore, the used data will be,
\begin{equation}
   \mathcal{D}=\left\{\mathbf{I}, \{ \mathbf{f}_j, \mathbf{p}_j, \mathbf{r}_j,\vec{f_j}, \vec{o_j}\}^M_{j=1} \right\}.
\end{equation}

Here, the function of the proposed DragOSM, when the model receives an image $\mathbf{I}$ and polygons ${\mathcal{P}}$, can be described ideally as:
\begin{equation}
    {\mathcal{R}}, {\mathcal{F}} = \mathrm{DragOSM}(\mathbf{I}, {\mathcal{P}}).
\end{equation}

The variables used in Sec.\ref{overview}, Sec.\ref{train} and Sec.\ref{test} are summarized in Table~\ref{tab:variables}.

\begin{table}
    \centering
        \caption{Important variables and their definitions for Sec.\ref{train} and Sec.\ref{test}.}
    \begin{tabular}{lr}

\toprule
        Variables& \#Definitions\\
\midrule
         $\mathbf{I}$& The input image.\\
         $\vec{v}$& A replacement expression for vector offsets $\vec{f}, \vec{r}, \vec{o}$. \\
         $\vec{v}_e$& The encoded $\vec{v}$ used in training. \\
         $\vec{\alpha}$, $\beta$& Parameters for the encode-decode of $\vec{v}$ and $\vec{v}_e$.\\
         $\mathcal{P},\mathcal{R},\mathcal{F}$ & Polygonal label sets for the input $\mathbf{I}$. \\
         $\boldsymbol{\mathcal{P}},\boldsymbol{\mathcal{R}},\boldsymbol{\mathcal{F}}$ & Polygon matrices which padded from sets.\\
         $\boldsymbol{\mathcal{M}}$ & A mask for the padded polygon matrices. \\
         $\vec{P},\vec{R},\vec{F}$ &  Offset label sets for the input $\mathbf{I}$.\\
         $\vec{\mathbf{P}},\vec{\mathbf{R}},\vec{\mathbf{F}}$ &  Offset matrices which padded from the offset sets.\\
         $\mathbf{\hat{\{ \cdot \}}}$ & A prediction value for ${\{ \cdot\}}$.\\
         {$\boldsymbol{\theta}_r, \boldsymbol{\theta}_f$} & Model parameters for roof and footprint prediction.\\
         $q(A|B; \boldsymbol{\theta}, \mathbf{I})$&Get the offset matrix for location ${B}$ to ${A}$ under $\boldsymbol{\theta}, \mathbf{I}$. \\
         $\gamma$ & A decaying factor in denoising inference.\\
         $t$ & The discrete index of $\boldsymbol{\mathcal{P}}$ for denoising. \\
\bottomrule
    \end{tabular}
    \label{tab:variables}
\end{table}

\subsection{Structure of DragOSM}
\label{overview}

To obtain polygonal results that better represent building structures, we designed DragOSM with a key principle: it must maintain the vectorized polygon format for both input and output.
We achieve this in DragOSM by predicting a vector offset for each input polygon, a departure from previous segmentation-based correction fields~\cite{Zampieri_2018_ECCV}. Instead, we propose the \textit{alignment token}, a concept inspired by foundation models, to directly encode and decode this alignment information. 
Inspired by interactive models like SAM~\cite{sam}, SAM~2~\cite{sam2}, DINO-X~\cite{ren2024dinox},~\etc, this entire process is formulated as an interaction between the label and image modalities, which we ultimately instantiate  DragOSM upon the SAM framework.

As shown in Fig.\ref{fig:dragosm}, our model adopts a Vision Transformer (ViT)~\cite{vit} as the image encoder. The polygonal representations of building keypoints are first converted into masks in the pixel space, which are then processed by a polygon encoder to generate polygon embeddings. The polygon encoder is implemented as a lightweight convolutional network, primarily responsible for mapping the polygon masks into a high-dimensional feature space. 

Then, polygon embeddings serve as the positional embedding for both the supervision task token and the alignment tokens. These tokens, together with the polygon embeddings, are grouped and fed into a two-way transformer~\cite{sam} to interact with the image embedding. The supervision task token and alignment tokens function analogously to the object queries in models such as DETR~\cite{detr} and DINO~\cite{dino}, acting as a set of learnable parameters that are randomly initialized during training.

Finally, the supervision task tokens are decoded together with the image embeddings to represent specific parts of the building, \ie, roof, footprint, building, and input polygon location, thereby helping the model learn a better structural understanding of buildings during training. In contrast, the alignment tokens are passed through a simple feed-forward network (FFN) and, within the polygon regression module, are decoded as two offsets mentioned in Eq.\ref{eq:relationship}. In the learning process, the FFN is trained to learn the encoded alignment offsets, while in the inference stage, the output of the FFN will be decoded as alignment offsets. $\vec{f}, \vec{r}$ and $\vec{o}$ share a similar encode-decode system. The encoding process is defined as:
\begin{equation}
\label{eq:align_encoder}
    \vec{v}_e = \frac{\vec{v}-\vec{\alpha}}{\beta}, 
\end{equation}
and the decoding process:
\begin{equation}
\label{eq:align_decoder}
\vec{v} = \beta\vec{v}_e+\vec{\alpha},
\end{equation}
where $\vec{v}_e$ and $\vec{v}$ are used to represent the encoded and decoded  $\vec{f}, \vec{r}, \vec{o}$, and $\vec{\alpha}$ is a hypothetical mean center. $\beta$ is a scale factor that normalizes the alignment offsets, which are measured in real-world pixel units, into a value range suitable for model learning.

The final offsets are combined with the input polygon keypoints to regress the final updated roof and footprint polygons in inference. 
As shown in Fig.\ref{fig:DragOSM_inference}, the DragOSM provides two core functions: the first is given a coarse polygon $\mathbf{p}_{t}$, DragOSM matches it to a closer location $\mathbf{p}_{t+1}$ with the footprint on the input images.
Another function is given a polygon $\mathbf{p}_{t+1}$ closer to footprint, DragOSM aligns it to a closer location $\hat{\mathbf{r}}$. 

Once the input image $\mathbf{I}$ is fixed, then the relationship between the roof and footprint is fixed, which means $\vec{o}$ is fixed. The certainty makes the estimation of the footprint-to-roof simpler than that of OSM-to-footprint. As a result, the OSM-to-footprint aligning process is iterative, while the footprint-to-roof aligning process is a one-step process. 

In the following sections, a more detailed description of the training procedure and inference process is provided for DragOSM.

\subsection{Hypothesis of Noise in Training}
\label{train}
In real-world scenarios, the positional discrepancies of historical labels relative to updated imagery are complex and stochastic. If a model is trained only on the fixed set of misaligned labels provided in a dataset, it operates under a closed-set assumption. This makes it difficult for the model to learn and generalize to the full, random distribution of label offsets found in reality.

As a result, DragOSM models the relationship between the original OSM labels and the true building positions in a given image as a Gaussian distribution. Accordingly, during training, Gaussian perturbations are added to the ground-truth labels to simulate positional noise, and the ground-truth footprints are finally applied to learn the fixed relationship between roofs and footprints.

More specifically, although the dataset $\mathcal{D}$ provides misaligned OSM polygon positions, these polygons are static. To enhance the robustness of our model, we dynamically add Gaussian noise to the ground truth footprints during training to simulate the displacement commonly observed in OSM data. While similar noise injection strategies have been used in previous near-nadir label correction algorithms~\cite{Zampieri_2018_ECCV, Girard_2018_ACCV}, those methods typically apply fixed offsets to the input polygons, resulting in a static misalignment relative to the building in the image. In contrast, our approach applies dynamically generated noise at each training iteration, leading to a more realistic and diverse simulation of annotation errors. By learning to denoise these perturbed labels, DragOSM acquires the capability to accurately update and the locations of correct building annotations. 

We choose $\vec{f_j}, \vec{o_j}$ from Eq.\ref{eq:relationship} as intermediate variables for positional correction based on a clear rationale: since footprints represent the ground-level outlines of buildings, their geographic positions are inherently more stable than roofs. Furthermore, for any given image, the spatial relationship between a building's footprint and roof is fixed, providing a reliable geometric relationship for correction.

We now detail the training procedure. During training, the input $m$ labels are randomly sampled to obtain $m'$ instances for each training iteration to better simulate the complexity of real-world labels. 
For $\mathcal{F} = \{ \mathbf{f}_j \}_{j=1}^{m'}$, each polygon is expanded into a matrix $\boldsymbol{\mathcal{F}}$ of shape $m' \times l \times 2$, where $l$ denotes the maximum number of keypoints among all polygons. Meanwhile, the sampled offset sets $\vec{F}$ and $\vec{O}$ are likewise extended into matrices $\vec{\mathbf{F}}$ and $\vec{\mathbf{O}}$ in the same shape. 
We use a binary matrix $\boldsymbol{\mathcal{M}}$ to perform a Hadamard product with $\boldsymbol{\mathcal{F}}$, allowing us to distinguish actual keypoints from padded entries introduced during the extension process. Specifically, elements corresponding to valid keypoints are set to 1 in $\boldsymbol{\mathcal{M}}$, while padded positions are set to 0. 

Subsequently, an additive Gaussian noise $\vec n \sim \mathcal{N}(\mathbf{0}, \sigma^2 \boldsymbol{I})$ is independently applied to each keypoint in $\boldsymbol{\mathcal{F}}$ to simulate annotation perturbations, where $\boldsymbol{I}$ is the identity matrix and $\sigma$ is a fixed standard deviation.
This noise injection process can thus be formulated as:
\begin{equation}
\label{eq:add_noise}
    \tilde{\boldsymbol{\mathcal{P}}}=(\boldsymbol{\mathcal{F}} - \vec{\mathbf{N}}) \odot \boldsymbol{\mathcal{M}}.
\end{equation}
In Eq.\ref{eq:add_noise}, we use the generated Gaussian noise field $\vec{\mathbf{N}}$ to replace $\vec{\mathbf{F}}$ as the system positional error that maps the ground truth polygons $\boldsymbol{\mathcal{F}}$ to their corresponding noised OSM positions $\tilde{\boldsymbol{\mathcal{P}}}$. 
Note that, although the dataset provides a fixed OSM polygon ${\boldsymbol{\mathcal{P}}}$, we use a dynamic $\tilde{\boldsymbol{\mathcal{P}}}$ to train our model. Therefore, the denoising learning process of DragOSM can be formulated as:
\begin{equation}
\label{eq:o2f}
\hat{\vec{\mathbf{F}}}=q(\boldsymbol{\mathcal{F}}|\tilde{\boldsymbol{\mathcal{P}}}; \boldsymbol{\theta}_f, \mathbf{I}),
\end{equation}
where $\boldsymbol{\theta}_f$ is model parameters related to foot alignment token, $\hat{\vec{\mathbf{F}}}$ is the model predicted value for ${\vec{\mathbf{F}}}$. In training, the ${\vec{\mathbf{F}}=\vec{\mathbf{N}}}$. 


Furthermore, since only $\vec{f}_j$ and $\vec{o}_j$ are used as correction vectors in Eq.\ref{eq:relationship}, and $\vec{o}_j$ is invariant to the noise in the OSM polygons. Here, we directly use ground truth $\boldsymbol{\mathcal{F}}$ as polygonal inputs, and the roof offset state transition matrix is given by:
\begin{equation}
\label{eq:f2r}
\hat{\vec{\mathbf{O}}}=q(\boldsymbol{\mathcal{R}}|\boldsymbol{\mathcal{F}};\boldsymbol{\theta}_r, \mathbf{I}),
\end{equation}
where $\boldsymbol{\theta}_r$ is model parameters related to roof alignment token, and $\hat{\vec{\mathbf{O}}}$ is the prediction for $\vec{\mathbf{O}}$.

During the noise-augmented training process, since the added Gaussian noise has a mean of $\mathbf{0}$, which effectively simulates a wide range of displacement magnitudes, only a single denoising step is performed in each training iteration. In contrast, during inference, multi-step denoising is applied to progressively refine the label positions.

Therefore, the overall training loss is defined as:
\begin{equation}
 \mathcal{L} = \sum^4_{k=1}\mathcal{L}_{s_k} + \gamma\times(\mathcal{L}_f+\mathcal{L}_o),
 \label{eq:loss}
\end{equation}
where $\gamma$ is a scaling factor, $\mathcal{L}_{s_k}$ denotes the CrossEntropy Loss~\cite{celoss} for the supervision tasks, and $\mathcal{L}_f$ and $\mathcal{L}_o$ represent the Smooth L1 Loss~\cite{fastrcnn} for related alignment offset learning.

\subsection{Denoising Process in Inference}
\label{test}

Due to the complexity of the positional discrepancies in historical labels, especially in cases with large errors, a simple one-step denoising process is often insufficient for correction. 
More critically, in off-nadir scenarios, a model must produce at least two distinct outputs (roof and footprint) to correctly represent a building. 
Therefore, we designed a multi-step, iterative inference strategy to address these challenging misalignments. 
Meanwhile, in Sec.\ref{ideal_dragosm}, we provide a heuristic analysis for the inference of DragOSM, and in Sec.\ref{tta}, we design inference-level strategies that achieve higher accuracy using the same number of denoising steps as the plain inference.

\begin{figure}
    \centering
    \includegraphics[width=1\linewidth]{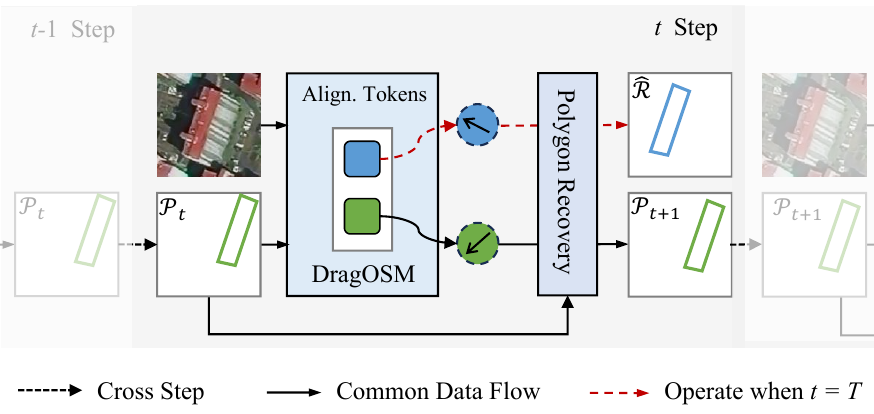}
    \caption{During inference, DragOSM iteratively adjusts the location of each footprint for every step. When reaching the final step $t=T$, the model will directly predict a roof location in one step. }
    \label{fig:DragOSM_inference}
\end{figure}

DragOSM’s multi-step denoising process treats denoising as the accumulation of positional noise derivatives, utilizing the property that derivatives of Gaussian processes remain Gaussian. In other words, the inference process effectively acts as a differentiator for positional noise. As Fig.\ref{fig:DragOSM_inference}, DragOSM obtains the final roof and footprint labels by iteratively updating the input polygons through a series of stepwise corrections.

To be specific, the inference ability of DragOSM is acquired from the training process as Eq.\ref{eq:o2f} and Eq.\ref{eq:f2r}. For an input pair $(\mathbf{I}, \boldsymbol{\mathcal{P}})$, we need to extract the footprint locations first. Here, we define a discrete sequence $\boldsymbol{\mathcal{P}}_0, ..., \boldsymbol{\mathcal{P}}_T$ and use $t$ to define a state in the sequence
, where $\boldsymbol{\mathcal{P}}_0$ represents the input state of $\boldsymbol{\mathcal{P}}$ and $\boldsymbol{\mathcal{P}}_T$ represents the final estimated location of the footprint $\hat{\boldsymbol{\mathcal{F}}}$. 
A decoding sequence for predicting footprints can be described as:
\begin{equation}
\label{eq:denoise_step}
\boldsymbol{\mathcal{P}}_t=\boldsymbol{\mathcal{P}}_{t-1}+a_t\cdot q(\boldsymbol{\mathcal{P}}_t|\boldsymbol{\mathcal{P}}_{t-1}; \boldsymbol{\theta}_f, \mathbf{I}),
\end{equation}
where $\{a_t\}_{t=1}^T$ is an real-valued sequence set indexed by the denoising step $t$ and serves as a scaling factor. For notational convenience in the remainder of this work, we set the scaling sequence to an exponential form:
\begin{equation}
a_t = \delta^{\,t-1},
\end{equation}
where  $\delta$ is a constant.

Therefore, the relationship between $\hat{\boldsymbol{\mathcal{F}}}$ and the input $\boldsymbol{\mathcal{P}}$ is given by:
\begin{equation}
\label{eq:o2f_test}
    \hat{\boldsymbol{\mathcal{F}}} = \boldsymbol{\mathcal{P}} + \sum_{t=1}^T\delta^{t-1} q(\boldsymbol{\mathcal{P}}_t|\boldsymbol{\mathcal{P}}_{t-1}; \boldsymbol{\theta}_f, \mathbf{I}). 
\end{equation}
In Sec.\ref{ablations}, we find that the most important factors affecting inference performance are the $T$ and $\delta$, and we will go further to understand the power of DragOSM’s denoising by analyzing the series $\sum_{t=1}^T\delta^{t-1}$ in Sec.\ref{dis_inference}, and a theoretical analysis is provided in Sec.\ref{ideal_dragosm}. 

Following this, with the prediction of $\hat{\boldsymbol{\mathcal{F}}}$, the relationship between roof and footprints is fixed once $\mathbf{I}$ is given. Applying Eq.\ref{eq:f2r}, we have the estimated roof location $\hat{\boldsymbol{\mathcal{R}}}$ in one step:
\begin{equation}
\label{eq:f2r_test}
    \hat{\boldsymbol{\mathcal{R}}} = \hat{\boldsymbol{\mathcal{F}}}+q(\hat{{\boldsymbol{\mathcal{R}}}}|\hat{\boldsymbol{\mathcal{F}}};\boldsymbol{\theta}_r, \mathbf{I}). 
\end{equation}

\begin{figure}
    \centering
    \includegraphics[width=1\linewidth]{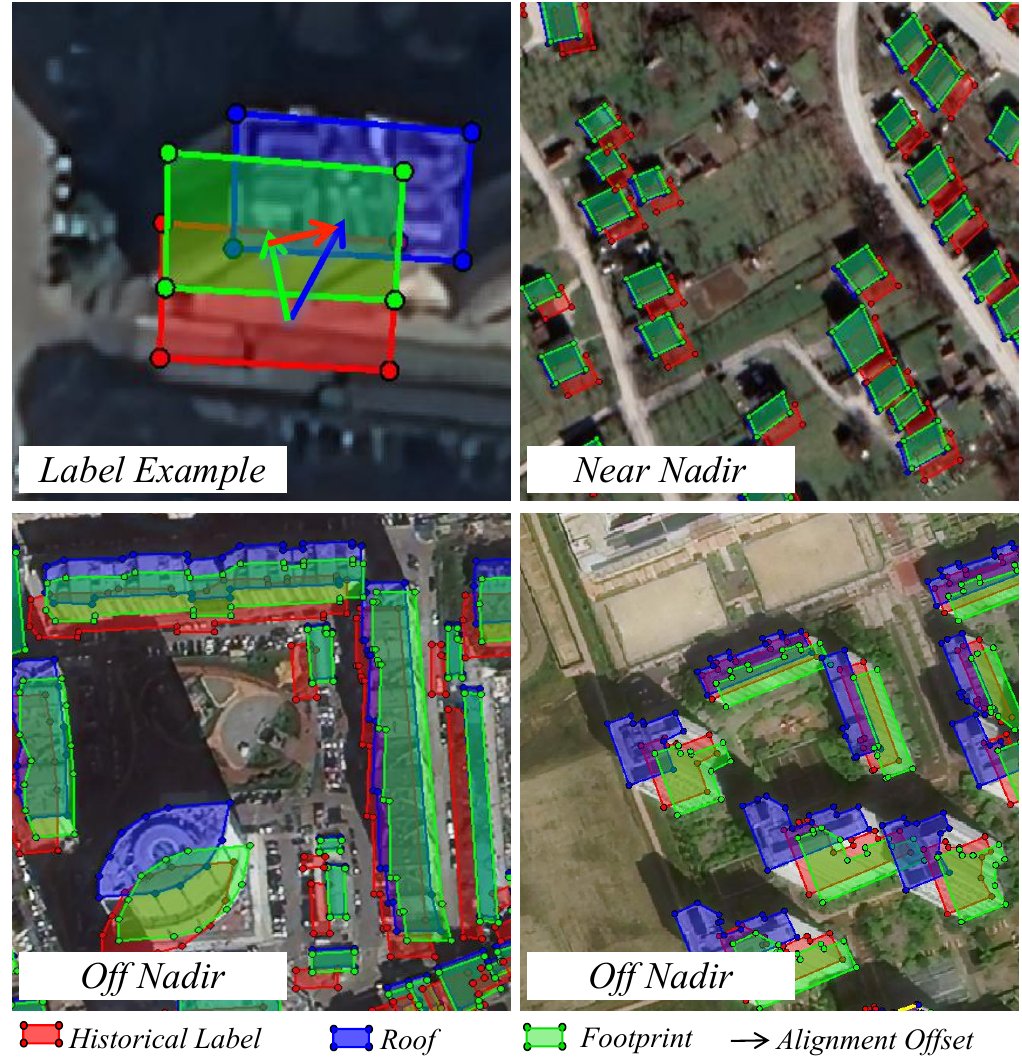}
    \caption{Label example and annotated samples in ReBO geolocated in different cities and camera angles. }
    \label{fig:data_example}
\end{figure}

From Eq.\ref{eq:o2f_test} and Eq.\ref{eq:f2r_test}, the label-in-label-out nature of DragOSM becomes more apparent, and its abilities can be decoupled into two parts: given polygons matching footprints, and given footprints matching roofs. 

Since the input $\boldsymbol{\mathcal{P}}$ can be arbitrary, this decoupled architecture provides remarkable versatility.
This allows DragOSM to do more than just denoise labels in near- and off-nadir images. Specifically, for off-nadir imagery, it can correct the projection parallax to find $\boldsymbol{\mathcal{F}}$ when given $\boldsymbol{\mathcal{R}}$ as input, or conversely, infer roof locations $\boldsymbol{\mathcal{R}}$ when given footprints $\boldsymbol{\mathcal{F}}$. 
Furthermore, the relative building height can be calculated simultaneously from the length of offset $\vec{o}$.

\section{Experiments}
This section presents the validation of our method. We begin by outlining the experimental setup, including the datasets (Sec.\ref{sec:dataset}), implementation details (Sec.\ref{details}), and baseline methods (Sec.\ref{baseline_method}). We then describe the evaluation protocols (Sec.\ref{protocals}), followed by a detailed analysis and visualization of the results in Sec.\ref{sec:main_result}.

\subsection{Datasets}
\label{sec:dataset}
\textbf{ReBO.} We built ReBO, a new dataset for evaluating alignment-based building polygon extraction methods. ReBO contains paired remote sensing imagery and historical building labels, where the imagery is sourced from Google Earth\footnote{{https://earth.google.com/}} and the historical labels are derived from OpenStreetMap. Specifically, the remote sensing imagery boasts a high spatial resolution of 0.5 m, ensuring fine-grained details of building structures are captured. The image collection includes both near-nadir and off-nadir views and encompasses a vast geographic distribution, spanning 41 cities across 22 countries on 6 continents. This wide coverage presents a highly diverse set of building styles, posing a significant challenge to the generalization capabilities of extraction methods.


To generate ground truth from historical labels, expert annotators manually align the initial polygons with the building outlines in the corresponding imagery. This process produces accurate footprint and roof polygons. The displacement offsets among original labels, roof polygons, and footprint polygons are simultaneously calculated and stored as key attributes. The trio of footprint polygons, roof polygons, and offset vectors forms the ground truth in ReBO. Fig.\ref{fig:data_example} provides representative visualizations of annotated samples.



The complete ReBO dataset comprises a total of 179,265 building instances across 5,473 images, all sized at $512\times512$ pixels. To facilitate supervised training, the dataset is partitioned into two standard subsets: a training set (5,003 images, 162,170 instances) and a test set (470 images, 17,101 instances). The partitioning is performed at the city level: the training split contains 35 cities, while the test split covers 8 cities. Notably, two cities appear in both the training and test splits, though with strictly non-overlapping geographic coverage. A summary of dataset statistics is provided in Table~\ref{tab:dataset_counts}, with further geospatial details available on our project homepage\footnote{https://github.com/likaiucas/DragOSM/}.

\begin{table}[t]
  \caption{Statistics of the proposed ReBO dataset partition.}
  \label{tab:dataset_counts}
  \centering
    \begin{tabular}{llrr}
      \toprule
      {Split} & {Viewing Angle} & {\# of Images} & {\# of Buildings} \\
      \midrule
      \multirow{2}{*}{Train}
        & Near-nadir & 1,935 & 60,411 \\
        & Off-nadir  & 3,068 & 101,759 \\
      \midrule
      \multirow{2}{*}{Test}
        & Near-nadir &   150 &  4,259 \\
        & Off-nadir  &   320 & 12,842 \\
      \midrule
        Total &--& 5,473 & 179,265 \\
      \bottomrule
    \end{tabular}
\end{table}

\textbf{BONAI \& OmniCity.}
BONAI~\cite{BONAI} and OmniCity~\cite{omnicity} are two open-source datasets designed for building footprint detection, providing annotations for each building that include its roof polygon, footprint polygon, and the roof-to-footprint offset. However, unlike our ReBO dataset, they lack annotations related to historical labels, such as OSM labels and the corresponding OSM-to-footprint or OSM-to-roof offsets. Consequently, the applicability of these datasets is limited to studies focused on direct roof-to-footprint and footprint-to-roof transformations.

The BONAI dataset consists of 3,300 remote sensing images (1024x1024 pixels) with spatial resolutions of 0.3m and 0.6m, encompassing a total of 268,958 building instances. The dataset is divided into a training set of 3,000 images and a testing set of 300 images.

The OmniCity dataset provides a multi-view perspective on urban buildings, offering five different viewing angles for the same urban areas. For our experiments, we use the OmniCity-view3 subset, which is widely adopted for footprint extraction research on off-nadir images. It contains 191,470 buildings distributed across 17,092 training and 4,929 testing images, each with a shape of 512x512 pixels.

\subsection{Implementation Details}
\label{details}
Following the pipeline in Fig.\ref{fig:dragosm}, we use ViT-base~\cite{vit} initialized on SAM~\cite{sam} as the backbone. All the models are trained with a batch size of 8 on 8 NVIDIA RTX 3090 GPUs (with 24GB RAM). 
We use 48 epochs for training, with a step learning rate of 0.02, which warms up from 0 in 500 iterations and decays by a factor of 0.1 at the 32nd and 44th epochs. The Stochastic Gradient Descent (SGD) with a weight decay of 0.0001 and momentum of 0.9 is used in all experiments. 
$\gamma$ in Eq.\ref{eq:loss} is set to 0.1 to balance the loss ranges between supervision tasks and alignment tasks.
During decoding, we set $\delta=1$ and $T=5$.  
All models are built in PyTorch, except MapAlignment~\cite {Zampieri_2018_ECCV}, whose released project was written in TensorFlow.

\subsection{Competitors}
\label{baseline_method}
DragOSM is designed to concurrently extract both roof and footprint polygons of buildings, and is suitable for use with both near-nadir and off-nadir imagery. To evaluate its performance, we select four distinct categories of methods for comparison. Specifically, the first category comprises common segmentation-based building extraction methods, whose architectures can be easily adapted to accommodate the extraction of both building roofs and footprints. We select three instance segmentation models (Mask RCNN~\cite{maskrcnn}, Cascade RCNN~\cite{cascadercnn}, and HTC~\cite{htc}) and two semantic segmentation models (UNet~\cite{unet2015} and HRNet~\cite{wang2020hrnet}). These models are widely adopted as backbone architectures for remote sensing image interpretation. In the second category, we choose LOFT~\cite{BONAI} and MLS-BRN~\cite{MLS-BRN}, both of which can simultaneously predict masks for building roofs and footprints. These methods incorporate sophisticated designs to address challenges inherent in off-nadir images. The third category includes specialized models, such as OBM~\cite{obm} and PolyFootNet~\cite{li2024polyfootnet}, that are capable of accepting prompts (\emph{e.g.}, bounding boxes marking roof locations) to perform building extraction. This configuration allows for a fairer comparison with our DragOSM, which also utilizes historical labels as additional input. Finally, to assess the advantages of our alignment framework in building polygon extraction, we compare DragOSM against MapAlignment~\cite{Zampieri_2018_ECCV, Girard_2018_ACCV, Girard_2019_IGARSS}, a method specifically developed to align historical labels with building roofs in near-nadir imagery.

\subsection{Evaluation Protocols}
\label{protocals}
Two categories of quantitative metrics are employed to evaluate building extraction performance. The first comprises mask-based metrics such as Intersection over Union (IoU), F1-score (F1), precision, and recall, as most competitors are only capable of predicting both roof and footprint masks. To enable a comprehensive evaluation, we introduce two macro-averaged metrics. \textit{Macro F1} (MF), computed as the mean of the F1 scores for roofs and footprints, and \textit{Macro IoU} (MI), representing the mean of their IoU scores. Converting these masks into polygons using a custom post-processing algorithm could introduce bias, since the post-processing step itself significantly influences accuracy. To ensure a fair comparison, we rasterize DragOSM's vector outputs into masks for evaluation against these methods. 

To rigorously evaluate the positional discrepancy between the predicted and ground-truth roofs and footprints, we adopt the Endpoint Error (EPE) metric. EPE measures localization accuracy by calculating the Euclidean distance between the centroid of a predicted polygon and the centroid of its corresponding ground-truth label.
The ability of DragOSM to estimate relative height for buildings is evaluated by Length Error ($\mathrm{LE}$), which is given by,
\begin{equation}
    \mathrm{LE} = \left|\left||\vec{o}||_2 - ||\hat{\vec{o}}||_2\right|\right|_2, 
\end{equation}
where $\vec{o}$ is the same as that in Eq.\ref{eq:relationship}.

In the evaluations, we use aLE to represent the average value of LE among all buildings in a dataset. Because the calculation of aLE and EPE requires a clear one-to-one correspondence between predictions and the ground truth, we could only evaluate OBM, PolyFootNet, and our own method using these metrics.

\subsection{Main Results}
\label{sec:main_result}
We conduct experiments on historical label-based extraction of building roof and footprint polygons using the ReBO dataset. For tasks involving footprint extraction from roof polygons (or vice versa) and building relative height estimation, we employ the BONAI and OmniCity datasets.

\begin{table*}[t]
  \footnotesize              
  \caption{Results (in \%) of historical label-based building polygon extraction on the ReBO dataset.}
  \label{tab:main_results}
  \centering
  \resizebox{\textwidth}{!}{%
  \begin{tabular}{
    @{}l
    *{5}{S[table-format=2.2]}
    *{5}{S[table-format=2.2]}
    *{3}{S[table-format=2.2]}@{}
  }
  \toprule
 \multirow{2}{*}{Model} &
  \multicolumn{5}{c}{{Roof}} &
  \multicolumn{5}{c}{{Footprint}} &
  \multicolumn{3}{c}{{Overall}} \\ 
  \cmidrule(lr){2-6}\cmidrule(lr){7-11}\cmidrule(l){12-14}
   & 
   {F1} & {Prec.} & {Rec.} & {IoU} & {EPE$\downarrow$} &
   {F1} & {Prec.} & {Rec.} & {IoU} & {EPE$\downarrow$} &
   {MF$\uparrow$} & {MI$\uparrow$}& {aLE$\downarrow$} \\
  \midrule
     Mask~R-CNN~\cite{maskrcnn}      & {76.31} & {68.73} & {89.11} & {62.81}& {-}& {72.35} & {65.22} & {84.21} & {55.65} &{-} & {-} & {-} & {-}\\
     Cas.~M.~R-CNN~\cite{cascadercnn}   & {77.53} & {70.56} & {89.16} & {64.55}& {-}& {72.54} & {65.66} & {84.10} & {56.10} &{-}& {-} & {-}& {-}\\
     HTC~\cite{htc}             & {77.11} & {69.65} & {89.66} & {64.01}& {-} & {71.42} & {65.96} & {80.46} & {57.08} &{-}& {-} & {-}& {-}\\
     U-Net~\cite{unet2015}           & {76.31} & {87.74} & {67.52} & {61.70}&{-} & {61.69} & {77.18} & {51.38} & {44.60} &{-}& {-} & {-}& {-} \\
     HRNet~\cite{wang2020hrnet}           & {83.44} & {85.95} & {81.08} & {71.59}&{-} & {79.58} & {80.52} & {78.36} & {65.87} &{-}& {-} & {-} & {-}\\
   MapAlign.~\cite{Zampieri_2018_ECCV}      & {64.91} & {65.00} & {64.86} & {50.36} &{13.77} & {-} & {-} & {-} & {-} &{-} & {-} & {-}& {-}\\
    MLS-BRN~\cite{MLS-BRN}        & {59.56}& {79.25} & {51.55} & {43.95}&{-} & {55.99} & {47.39} & {77.53} & {40.31} &{-}& {57.77} & {42.13} & {-}\\
    LOFT~\cite{BONAI}            & {77.55} & {67.35} & \textbf{94.31} & {64.60}&{-}& {74.12}&	{63.48}&	\textbf{92.06}&	{60.07} &{-}& {75.83}&{62.33}& {-}
 \\
  \midrule
    OBM~\cite{obm}             & \underline{88.13} & \underline{90.68} & {86.04} & \underline{79.13} &{2.42}& {80.66} & \underline{82.68} & {79.02} & {68.73} &{4.61}& \underline{84.40} & \underline{73.93} & {2.89}\\
    PolyFootNet~\cite{li2024polyfootnet}     & {86.52} & {84.71} & {88.88} & {76.60}&{3.12} & \underline{81.30} & {79.49} & {83.54} & {69.16} &{4.84}& {83.91} & {72.88}& {2.67} \\
  \midrule
  Ours (One Step) & {85.71}&	{86.97}&	{84.54}&	{75.99}&{5.74}&	{80.71}&	{81.23}&	{80.23}&	\underline{69.29}&{7.08}&		{83.21}&	{72.64}& {2.78}\\
   Ours (Multi-Step) & \textbf{91.36} & \textbf{92.33} & \underline{90.45} & \textbf{85.24}&\textbf{2.39}
           & \textbf{91.74} & \textbf{92.03} & \underline{91.47} & \textbf{85.24}
           &\textbf{2.26}& \textbf{91.55} & \textbf{84.83} & \textbf{2.04}\\
  \bottomrule
  \end{tabular}}%
  \begin{tablenotes}[flushleft]  
      \footnotesize
\item \textbf{Bold} highlights the best score; \underline{underline} marks the second best.
$\uparrow$: higher is better. $\downarrow$: lower is better. \textit{Prec.} and \textit{ Rec.} represent Mask Precision and Recall. The reported multi-step DragOSM performance was achieved with a single, fixed $\gamma=1, T=5$. 

    \end{tablenotes}
\end{table*}

\textbf{Historical Label-based Building Polygon Extraction.} Table~\ref{tab:main_results} summarizes the quantitative performance of all baseline and proposed methods on the ReBO dataset. 
Early methods for roof and footprint extraction, typically based on instance or semantic segmentation, struggle with off-nadir images because they overlook a critical issue: the ambiguous semantic boundary between the facade and the partially occluded footprint. This inherent limitation prevents the direct and accurate segmentation of the footprint, leading to poor performance. The performance disparity is evident in the large gap between roof and footprint metrics; \eg, with the most complex encoder architecture among the models, HRNet delivers the best overall performance for roof and footprint extraction. However, it still exhibits a notable F1-score gap of 3.86\% between its roof and footprint results.
Although MapAlignment can perform simple positional corrections for historical labels, it fails to resolve the displacement between roofs and footprints in off-nadir imagery. This is because the displacement is non-uniform across the image, varying with the height of each building. Consequently, methods that rely on a single, global correction, similar to predicting an optical flow field, are rendered ineffective. This limitation is reflected in its poor performance on roof extraction, achieving a Roof F1 score of only 64.91\%. 
LOFT employs a soft-NMS algorithm for post-processing, which retains a large number of low-confidence outputs. This strategy allows it to achieve the highest Roof and Footprint Recall scores (94.31\% and 92.06\%, respectively). However, this high recall comes at the cost of precision, resulting in a significant number of duplicate predictions and false detections.

OBM, PolyFootNet, and our DragOSM model represent a similar class of methods that leverage prior information (\eg, historical labels or prompts) as input. This approach yields a significant performance boost over traditional models; for instance, OBM achieves an impressive MF of 84.4\%, ranking second overall.

However, DragOSM demonstrates superior efficiency and capability. With just a single denoising step, our model already achieves performance comparable to OBM. Furthermore, after applying its multi-step denoising process, DragOSM establishes a new state-of-the-art, showing dominant performance across all key metrics, including relative height prediction (aLE 2.04), roof-footprint localization (EPE), and MF (91.55\%).

\begin{figure*}
    \centering
    \includegraphics[width=1\linewidth]{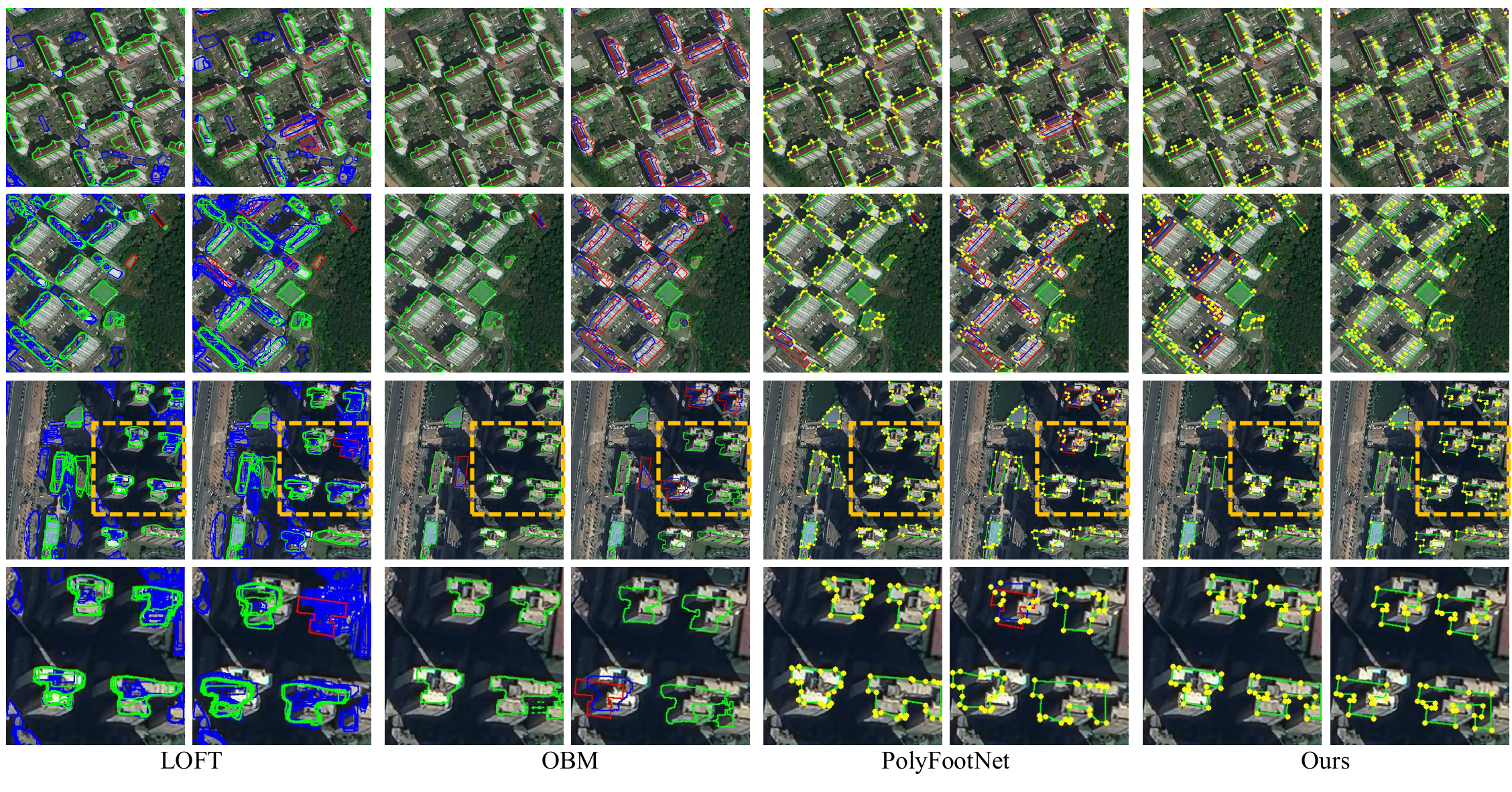}
    \caption{The results of the baselines and our method on the ReBO dataset off-nadir images. For each method, we present the roof predictions on the left and footprint results on the right. The green, blue, and red boundaries denote the TP, FP, and FN. The yellow nodes are keypoints of the predicted polygons. Three cases are provided, and the bottom row presents a magnified view of local details to better illustrate the subtle differences in model performance.}
    \label{fig:vis_off_nadir}
\end{figure*}

\begin{figure}
    \centering
    \includegraphics[width=1\linewidth]{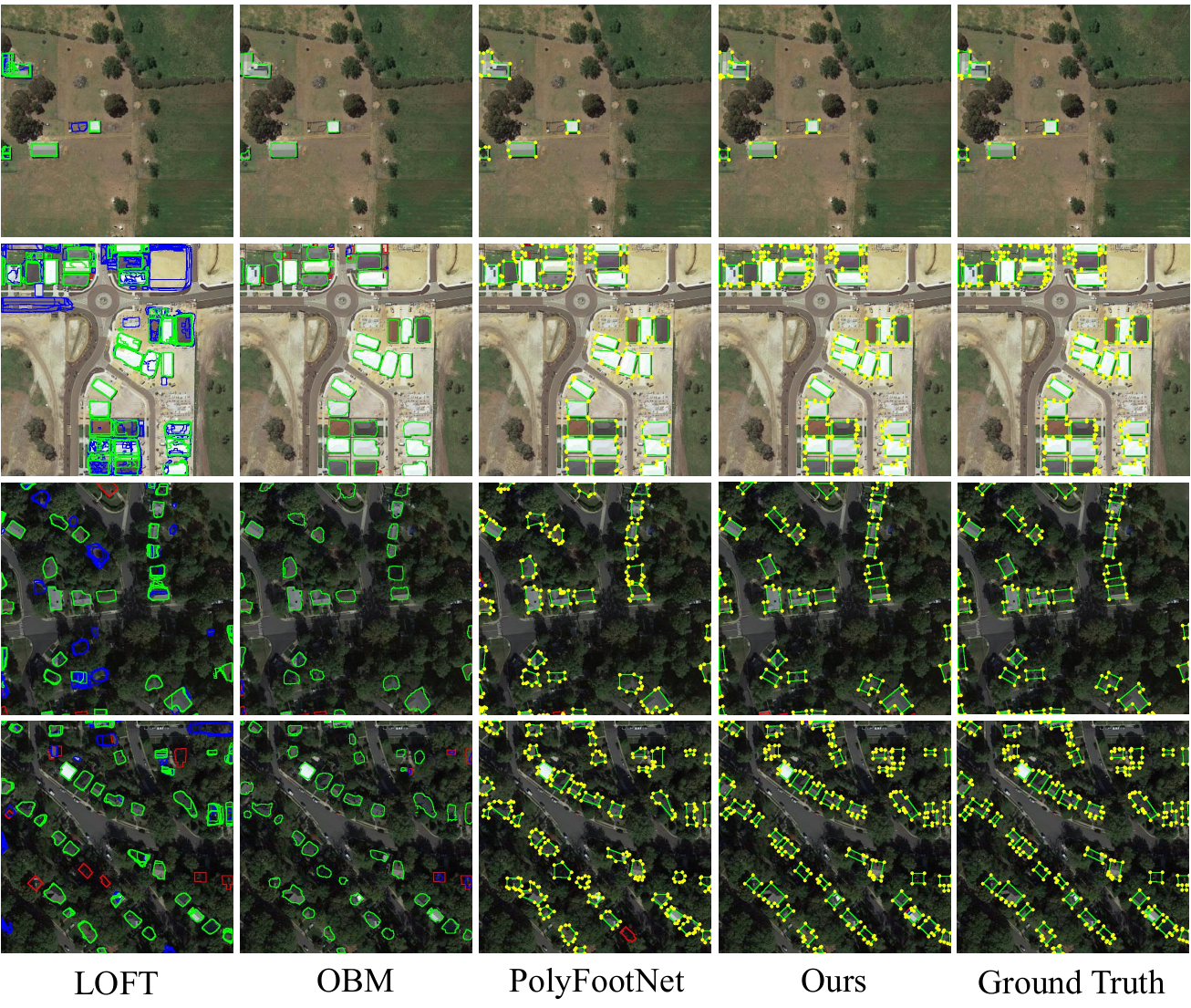}
    \caption{The visualized results of methods on the ReBO dataset near-nadir images. The green, blue, and red boundaries denote the TP, FP, and FN. The yellow nodes are keypoints of predicted polygons.}
    \label{fig:vis_near_nadir}
\end{figure}

\textbf{Visualization.} To better illustrate the differences between our method and the specialized off-nadir algorithms in Table~\ref{tab:main_results}, we have visualized several qualitative results in Fig.\ref{fig:vis_off_nadir} and Fig.\ref{fig:vis_near_nadir} for the understanding of the presented metrics. 
Although LOFT achieves extremely high Recall by using algorithms like soft-NMS, it also produces a large number of false and duplicate detections.
While OBM's reliance on controllable visual prompts leads to fewer false detections, its coarse mask boundaries for roofs and footprints prevent the high-quality vectorization of the results.
Similarly, although PolyFootNet can directly predict roof and footprint polygons by incorporating a keypoint prediction task, it is prone to significant errors. Its inaccurate estimation of the building's displacement often leads to incorrectly placed footprints. Furthermore, it can produce inconsistent edge and keypoint predictions for the same building. These issues of placement error and prediction instability are common to both PolyFootNet and OBM. These problems are commonly found in the roof and footprint extraction results of all three models (see Fig.\ref{fig:vis_off_nadir} and Fig.\ref{fig:vis_near_nadir}).

In comparison, our DragOSM model uses historical labels as input. By starting with these manually annotated labels, which provide a consistent and high-quality initial estimate of the building's shape, and combining this with a multi-step denoising inference process, DragOSM achieves the best results for both roofs and footprints. 

In addition to the findings above, a closer analysis of Fig.\ref{fig:vis_off_nadir} reveals a notable difference in the failure modes. The failure cases for OBM and PolyFootNet are primarily concentrated on footprint extraction, whereas DragOSM's errors are mainly in the roof predictions. We attribute this pattern to error accumulation in their respective two-stage processes: OBM and PolyFootNet first predict the roof and then derive the footprint, while DragOSM operates in reverse, first aligning the footprint and then deriving the roof.

\textbf{Bidirectional Roof-Footprint Alignment.}
Table~\ref{tab:merged_results} summarizes the roof and footprint alignment ability of DragOSM. Here, “w.” and “w/o.” indicate whether the corresponding dataset is used for training. 
Specifically, we exclusively use the roof and footprint data from the BONAI and OmniCity datasets, tasking the model to predict the corresponding footprint when given a roof, and vice versa.
For the roof-to-footprint alignment task, DragOSM is configured with $\delta=0.9$ and employs five denoising steps ($T=5$) for alignment. DragOSM performs favorably, ranking among the top contenders in several key metrics.

Overall, although DragOSM is not specifically designed for these two tasks, it consistently outperforms certain prompt-based models that have access to ground-truth location cues. However, conditioning on the roof or footprint position undermines the Gaussian distribution assumption of the data space established during DragOSM’s training process, potentially limiting the model’s ability to generalize to natural misalignments.

\begin{table*}
\small
\centering
    \caption{Results (in \%) of bidirectional roof-footprint alignment on BONAI and OmniCity datasets.}
    \label{tab:merged_results}
\scalebox{0.9}{
    \begin{tabular}{@{}llcrrrrrrrrrr@{}}
        \toprule
        \multirow{2}{*}{{Model}} & \multirow{2}{*}{{Dataset}} & \multirow{2}{*}{{Train}} & \multicolumn{5}{c}{{Footprint-to-roof alignment}} & \multicolumn{5}{c}{{Roof-to-footprint alignment}} \\
        \cmidrule(l){4-8}\cmidrule(l){9-13}
         & & & {F1$_r$} & {Prec.$_r$} & {Rec.$_r$} & {EPE$_r \downarrow$} & {a$LE \downarrow$} & {F1$_f$} & {Prec.$_f$} & {Rec.$_f$} & {EPE$_f$ $\downarrow$} & {a$LE \downarrow$} \\
        \midrule
        \multirow{2}{*}{PolyFootNet} & BONAI & w. & 85.09 & 85.11 & 85.68 & \underline{3.03} & 5.05 & 74.49 & 76.89 & 73.00 & \underline{6.78} & 5.05 \\
         & OmniCity & w. & \textbf{93.87} & \textbf{93.00} & \textbf{93.38} & \textbf{2.25} & \underline{4.65} & \textbf{88.42} & \textbf{90.06} & \textbf{87.01} & 6.84 & \textbf{4.65} \\
        \midrule
        \multirow{2}{*}{OBM} & BONAI & w. & 70.07 & 80.61 & 62.81 & 8.17 & 5.63 & 63.18 & 74.91 & 55.41 & 10.64 & 5.63 \\
         & OmniCity & w. & 88.37 & \underline{91.42} & 85.85 & 5.24 & 5.15 & \underline{86.03} & 90.17 & 82.52 & \textbf{6.20} & \underline{5.15} \\
        \midrule
        \multirow{4}{*}{Ours} & \multirow{2}{*}{BONAI} & w/o. & 73.49 & 74.85 & 73.48 & 9.44 & 8.76 & 68.06 & 69.30 & 67.28 & 10.96 & 12.56 \\
         & & w. & 83.04 & 84.81 & 81.77 & 4.94 & \textbf{3.91} & 74.32 & 82.88 & 79.26 & 8.67 & 6.26 \\
        \cmidrule(l){2-13}
         & \multirow{2}{*}{OmniCity} & w/o. & 88.59 & 88.23 & 89.06 & 6.87 & 6.32 & 83.38 & 85.50 & 81.47 & 10.68 & 7.25 \\
         & & w. & \underline{90.56} & 90.48 & \underline{90.74} & 6.18 & 4.75 & 85.12 & \underline{87.21} & \underline{83.22} & 9.75 & 6.19 \\
        \bottomrule
    \end{tabular}
     
    }
\vspace{2mm}

   \scriptsize{\textbf{Bold} highlights the best score; \underline{underline} marks the second best. $\uparrow$: higher is better. $\downarrow$: lower is better. }
\end{table*}
\section{Ablations}
\label{ablations}
In Sec.\ref{aligncoder}, we first determine the encoding and decoding parameter $\beta$ for the alignment tokens. Then, in Sec.\ref{alignstrat}, we explore different training strategies, including combinations of osm-to-footprint, osm-to-roof, and footprint-to-roof offsets.
In Sec.\ref{deltat}, we analyze the impact of step size settings during the denoising inference phase on the final results. Finally, Sec. \ref{sec:dataset_ablation} presents an ablation study on the dataset to separately evaluate the impact of off-nadir and near-nadir images on the final results.

\subsection{Parameters for Decode \& Encode Alignment Offsets}
\label{aligncoder}
In this part, we examine the encoding and decoding parameters for the proposed concept of the alignment token. In real-world scenarios, the ground-truth offset labels represent values in pixel space, often ranging from 1 to over 100, which are substantially larger than the intermediate values typically encountered during model training. To evaluate this impact, different combinations of the decoding parameters were tested.

\begin{table}[t]
  \centering
  \caption{Effect of $\delta$ in token coders on one-step overall accuracy.}
  \label{tab:delta_ablation}
  \setlength{\tabcolsep}{5pt}
  \renewcommand{\arraystretch}{1.15}

  \begin{tabular}{
llrr  
  }
    \toprule
    {{Roof $\delta$}} &
    {{Foot $\delta$}} &
    {{Macro F1}} &
    {{Macro IoU}} \\
    \midrule
    200 & 100 & 72.25 & 58.76 \\
    200 & 150 & 82.50 & 71.72 \\
    200 & 200 & \textbf{83.21} & \textbf{72.64}\\
    200 & 250 & 82.60 & 71.10 \\
    100 & 100 & 77.53 & 65.15 \\
    100 & 150 & 79.04 & 67.05 \\
    150 & 150 & 81.28 & 70.03 \\
    250 & 250 & 82.89 & 72.23 \\
    \bottomrule
  \end{tabular}
\end{table}

Specifically, different $\beta$ for decoding \& encoding both roof and footprint alignment tokens in Eq.\ref{eq:align_encoder} and Eq.\ref{eq:align_decoder} are studied by re-training with different $\beta$. 

In Table~\ref{tab:delta_ablation}, the model's performances are compared under one-step inference, and we found $\beta = 200$ to be a more suitable setting for both roof and footprint alignment. 

\subsection{Alignment Offset Combinations}
\label{alignstrat}
This section analyzes the impact of key design choices on model performance under a single-step inference setting. Motivated by the three possible alignment directions identified in Eq.\ref{eq:relationship}, we first investigate the performance of various alignment offset combinations. To do this, we retrain and compare multiple DragOSM variants, each with a unique setting. Additionally, we evaluate the effect of incorporating noise augmentation during training. The results of these comparisons are used to determine the most effective model configuration.

As shown in Table~\ref{tab:component_ablation}, Gaussian noise augmentation is beneficial for learning label alignment, and the combination using the $\vec{o}$-based alignment learning achieves better performance. This may be attributed to the inherent properties of $\vec{o}$: given an image, the relationship between the building roof and footprint, as well as the value of $\vec{o}$, is determined. In this setting, the model can more easily learn building-related information directly from the image, making the learning of $\vec{o}$ more straightforward compared to $\vec{f}$ and $\vec{r}$.

Based on these findings, DragOSM adopts the noise augmentation strategy and trains using the combination of $\vec{o}$ and $\vec{f}$.

\begin{table}[t]
  \centering
  \caption{Component ablation on training strategies.}
  \label{tab:component_ablation}
  \setlength{\tabcolsep}{3pt}        
  \renewcommand{\arraystretch}{1.15} 

  \begin{threeparttable}
  \begin{tabular}{
      c c c c
rr
  }
    \toprule
    $\vec{r}$ & $\vec{f}$ & $\vec{o}$ & {D.N.} &
    {{Macro F1}} & {{Macro IoU}} \\
    \midrule
    $\checkmark$ & $\checkmark$ &            &       $\checkmark$     & 76.41 & 63.68 \\
    $\checkmark$ &             & $\checkmark$ &       $\checkmark$    & 77.14 & 64.36 \\
                 & $\checkmark$ & $\checkmark$ &    $\checkmark$       & \textbf{83.21} & \textbf{72.64} \\
                 & $\checkmark$ & $\checkmark$ &  & 75.10 & 62.02 \\
    \bottomrule
  \end{tabular}

  \end{threeparttable}
\end{table}

\subsection{Stepsize for Denoise Inference}
\label{deltat}
During inference, we use $\{a_t\}_{t=1}^T$ to guide the model’s iterative correction process. Since an exponential sequence can simulate varying step sizes at different steps by flexibly setting its base, we set $a_t = \delta^{t-1}$.
To better illustrate the interplay between $\delta$ and $T$, we define:
\begin{equation}
\label{seriesum}
    E = \sum_{t=1}^{T} \delta^{t-1}=\frac{\delta-\delta^T}{\delta-1}, 
\end{equation}
which quantifies the ``energy'' accumulated throughout the denoising steps. 

From Table~\ref{tab:delta_analysis}, we notice that when the number of steps is small, the model’s final performance may be relatively insensitive to the specific values of $\delta$ and $T$. The results also indicate that the model tends to achieve similar performance when the cumulative energy $E$ is comparable. However, when $\delta$ becomes excessively large, the model’s performance appears to decline.

To further investigate this phenomenon, we conduct additional experiments and provide a comprehensive analysis of the inference process from multiple perspectives. 
By using a constant step length, we discover that directly setting a larger step size at the early stage allows for faster network fitting (see Sec.\ref{dis_inference}).

\begin{table}[t]
  \centering
  \caption{Effect of decay factor $\delta$ and step count $T$ on roof/footprint F1 and Macro F1.}
  \label{tab:delta_analysis}
  \setlength{\tabcolsep}{4pt}
  \renewcommand{\arraystretch}{1.15}
  \begin{tabular}{cccccc}
    \toprule
    $\delta$ & $T$ & $E$ & Roof F1 & Footprint F1 & MF \\
    \midrule
    \addlinespace[2pt]
    Any     &  1 & 1.00     & 84.54 & 80.23 & 82.39 \\
        \midrule
    0.1   &  5 & 1.11     & 84.89 & 80.85 & 82.87 \\
    0.3   &  5 & 1.42     & 85.91 & 82.61 & 84.26 \\
    0.5   &  5 & 1.93     & 87.43 & 85.09 & 86.26 \\
    0.9   &  5 & 4.09     & 90.16 & 90.75 & 90.46 \\
    1     &  5 & 5.00     & 90.46 & 91.47 & 90.96 \\
    1.1   &  5 & 6.10     & {90.62} & {91.95} & {91.29} \\
    1.2   &  5 & 7.44     & 90.57 & 91.60 & 91.09 \\
    1.3   &  5 & 9.04     & 90.33 & 90.87 & 90.60 \\
    \midrule
    \addlinespace[2pt]
    0.1   & 10 & 1.11     & 84.89 & 80.85 & 82.87 \\
    0.3   & 10 & 1.42     & 85.92 & 82.62 & 84.27 \\
    0.5   & 10 & 1.99     & 87.54 & 85.32 & 86.43 \\
    0.9   & 10 & 6.51    & 90.45 & 91.83 & 91.14 \\
    1     & 10 & 10.00    & 90.68 & 92.12 & 91.40 \\
    1.1   & 10 & 15.93    & 90.63 & 91.48 & 91.06 \\
    1.2   & 10 & 25.95    & 84.86 & 80.76 & 82.81 \\
    \bottomrule
  \end{tabular}
\end{table}

On the other hand, the inference speed of the model gradually decreases as the number of denoising steps increases. To quantify the impact of different $T$ on inference efficiency, we conducted five repeated experiments on the ReBO dataset and measured the average inference speed using the FPS (frames per second) metric under various $T$ settings (see Table~\ref{tab:fps_vs_T}).

\begin{table}[t]
  \centering
  \caption{Inference speed (FPS) versus number of denoising steps $T$.}
  \label{tab:fps_vs_T}
  \setlength{\tabcolsep}{6pt}        
  \renewcommand{\arraystretch}{1.1}   
  \begin{tabular}{crrrrrrr}
    \toprule
    $T$   & 1    & 2    & 4    & 5    & 8    & 10   & 30   \\
    \midrule
    FPS   & 1.43 & 1.10 & 0.91 & 0.69 & 0.55 & 0.48 & 0.10 \\
    \bottomrule
  \end{tabular}
\end{table}

\subsection{Impact of the Near-nadir and Off-nadir Images}
\label{sec:dataset_ablation}
The ReBO dataset was designed for a comprehensive evaluation of model performance across both off-nadir and near-nadir imagery. While we reported the overall results on the entire dataset in Sec.~\ref{sec:main_result}, these aggregate metrics do not fully reveal the specific impact of near-nadir versus off-nadir conditions on each model. To provide a more fine-grained analysis, the experiments in this subsection re-evaluate the models by splitting the dataset into distinct near-nadir and off-nadir subsets and assessing their performance on each.

As shown in Table~\ref{tab:model_performance_pro_no_mf}, traditional models attempt to directly extract roofs and footprints from the image's semantic information. However, the ambiguous semantic boundaries inherent in off-nadir imagery, which are more pronounced than in near-nadir views, complicate the model's ability to interpret pixel-level details. This leads to a significant performance gap between the extracted roof and footprint results. For example, on the off-nadir subset, UNet's Roof F1 is 13.7\% higher than its Footprint F1. While more complex traditional models like HRNet can improve overall performance by increasing their parameter count, the fundamental performance disparity between roof and footprint predictions persists.

To mitigate this, "expert" models such as OBM and PolyFootNet were introduced with improved building representations (predict \textit{roof + offset}) tailored for the off-nadir problem to enhance building understandings of models. Nevertheless, they still have deficiencies; as building displacement increases, so does the difficulty of offset prediction, and a performance gap between roof and footprint extraction remains. This is evidenced by PolyFootNet, which still exhibits an F1 gap of 4.19\% on the off-nadir subset.

In comparison, our model starts from historical labels, a strategy that bypasses the issue of pixel-level semantic ambiguity from the outset. Through its multi-step denoising process, DragOSM develops a significantly enhanced understanding of both projection and positional offsets. This leads to consistently accurate extraction results for both roofs and footprints, demonstrated by a minimal F1 gap of only 0.95\% and a much lower off-nadir aLE of 2.21.

\begin{table}
  \captionsetup{labelsep=newline, justification=centering} 
  \caption{Results (in \%) of different models on Off Nadir and Near Nadir scenes on ReBO test set.}
  \label{tab:model_performance_pro_no_mf}
  \centering
  \begin{tabular}{lcccccc}
    \toprule
     \multirow{2}{*}{Model} & \multicolumn{3}{c}{{Off Nadir}} & \multicolumn{3}{c}{{Near Nadir}} \\
    \cmidrule(r){2-4} \cmidrule(l){5-7}
    & aLE & F1$_r$& F1$_f$& aLE  & F1$_r$ & F1$_f$  \\
    \midrule
    UNet&{-}&76.63&62.93&{-}&75.37	&65.96\\
    HRNet           & {-}                      & 83.27                      & 78.78                      & {-}                      & 84.30                      & 81.98                      \\
    OBM             & 3.36                      & \underline{88.67}          & 78.57                      & \underline{1.36}          & \underline{89.44}          & \underline{86.84}          \\
    PolyFootNet     & 3.02                      & 86.55                      & 80.27                      & 1.57                      & 86.46                      & 84.28                      \\
    Ours(One Step)   & \underline{2.96}          & 84.71                      & \underline{80.52}          & 2.19                      & 87.85                      & 80.86                      \\
    Ours(Multi-Step) & \textbf{2.21}             & \textbf{90.22}             & \textbf{91.17}             & \textbf{1.29}             & \textbf{93.79}             & \textbf{92.94}             \\
    \bottomrule
  \end{tabular}

\end{table}

\section{Further Probing \& Discussions}
\label{discuss}
In Sec.\ref{dis_inference}, we examine in greater detail how to set the step size and the number of steps during inference from an engineering perspective. Then, Sec.\ref{ideal_dragosm} provides a heuristic explanation of the underlying principles of DragOSM. Finally, Sec.\ref{benefit} and Sec.\ref{limits_future} analyze the advantages and limitations of DragOSM, as well as its potential impact on future developments.

\subsection{Step Size Strategies and Inference Convergence}
\label{dis_inference}
In Sec.\ref{deltat}, we observed a potential relationship between the $\delta$ and $T$ parameters and the convergence behavior of the model during inference. To further investigate and analyze this relationship across different buildings, we visualize the scatter distributions of $E$ and Macro F1 under various $\delta$ values. This visualization provides an intuitive means to examine and validate the conclusions summarized previously.

\begin{figure}
    \centering
    \includegraphics[width=1\linewidth]{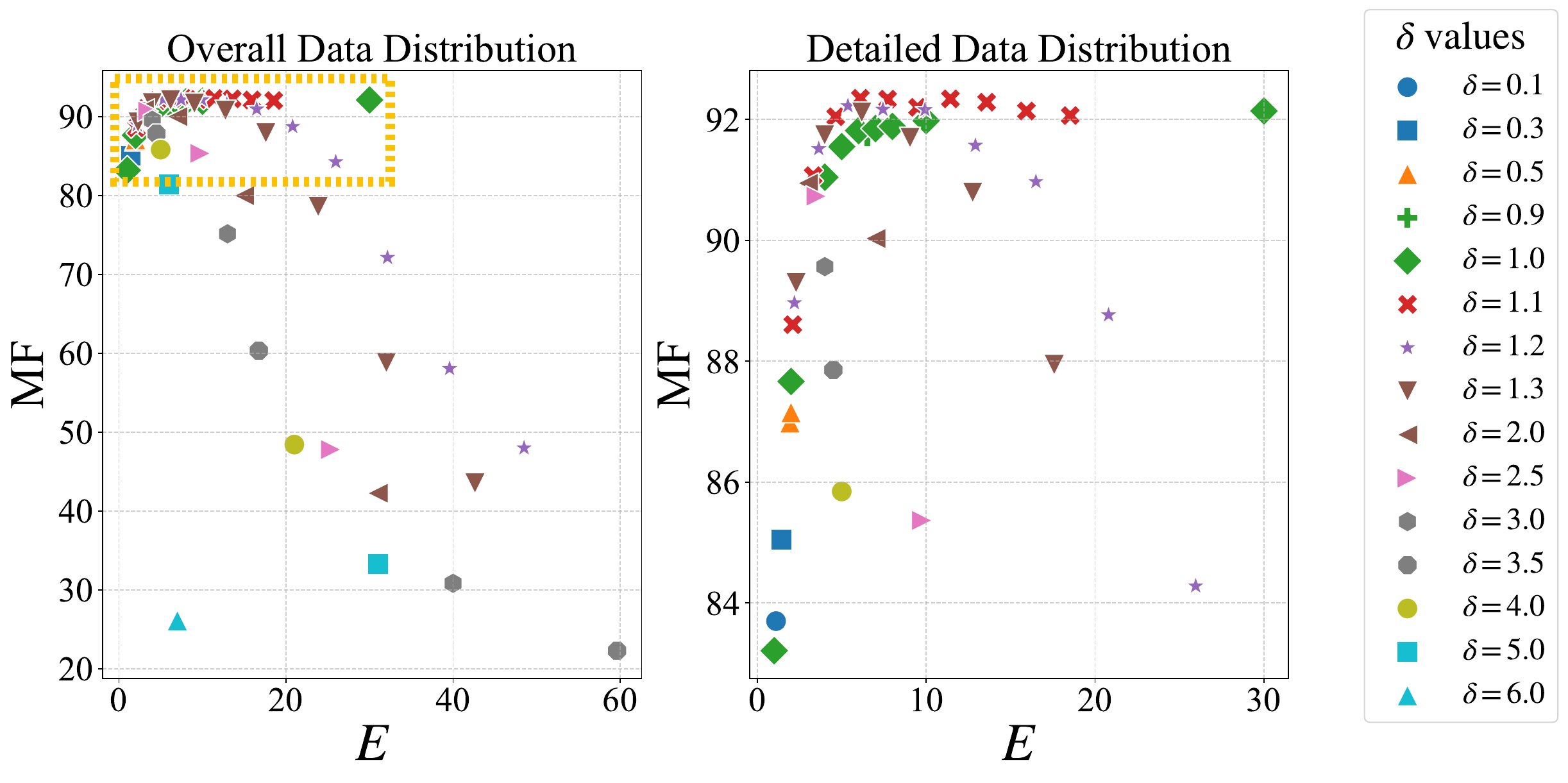}
    \caption{The relationship between energy $E$ and inference convergence. The graph on the left provides an overall $E-$MF distribution, while the graph on the right zooms into the convergence part in a more detailed view. }
    \label{fig:scatter}
\end{figure}

The Fig.\ref{fig:scatter} illustrates the relationship between model inference performance and $E$. Overall, given a fixed $\delta$, the model's performance is directly related to $E$ during the denoising process: before reaching optimal performance, models with similar $E$ values tend to achieve comparable results; \ie, a large $\delta$ could accelerate the convergence at early steps. However, as the number of denoising steps $t$ increases and the model approaches its optimal performance, the value of $\delta$ becomes a decisive factor in determining whether convergence is ultimately achieved. 

Therefore, analysis of inference convergence must return to examining the convergence behavior of the series $\sum_{t=1}^T a_t$ and $\sum_{t=1}^T q(\boldsymbol{\mathcal{P}}t \mid \boldsymbol{\mathcal{P}}{t-1}; \boldsymbol{\theta}_f, \mathbf{I})$ as defined in Eq.~\ref{eq:denoise_step}.

In our experiments, we set $a_t = \delta^{t-1}$. According to the geometric series summation in Eq.~\ref{seriesum}, it is evident that when $\delta < 1$, $\sum_{t=1}^T a_t$ converges, ensuring model convergence. For $\delta = 1$, although $\sum_{t=1}^T a_t = T$ diverges, our empirical results show that the model’s inference process still converges, indicating that DragOSM inherently possesses strong convergence properties. Furthermore, even when $\delta$ is set to a larger value and the divergence of $\sum_{t=1}^T a_t$ becomes more pronounced, we still observe convergence within certain intervals in our experiments. 

These findings indicate that the model indeed corrects label displacement errors by predicting a sequence of offsets. While a larger $\delta$ may reduce the robustness of the denoising process to some extent, appropriately setting $\delta$ and $T$ can provide beneficial length compensation for each predicted offset, thereby accelerating the convergence of the model. Specifically, some practical combinations include $(\delta, T) = (1, 5)$, $(1.3, 4)$, and $(2, 2)$.

This highlights a fundamental difference between our approach and diffusion models, despite DragOSM being inspired by the denoising strategies employed in diffusion models.
Traditional diffusion models for generations, such as DDPM~\cite{NEURIPS2020_ddpm}, introduce noise to image samples during training and employ KL divergence to learn the denoising process, enabling the recovery of meaningful images from pure noise distributions. This inference procedure typically requires dozens or even hundreds of iterative steps. When examining the same image under different training iterations with noise added, the resulting images are essentially pure noise and do not retain any meaningful distribution related to the original image content. \textbf{In contrast}, DragOSM introduces noise to the spatial positions of the labels rather than the image pixels themselves, and uses the Smooth-L1 Loss to learn the added noise. This distinction means that, for a given image and label, the added noise across different training iterations follows a clear pattern: the aggregated noisy label positions form a Gaussian distribution with mean $\mathbf{0}$, \ie, $\mathcal{N}(\mathbf{}{0}, \sigma^2 \boldsymbol{I})$. Importantly, the mean of this distribution corresponds exactly to the true label position. During inference, DragOSM corrects label positions by predicting a trajectory that follows this Gaussian distribution, typically requiring fewer than ten steps to achieve accurate alignment, which is significantly fewer than traditional diffusion models.

\subsection{Heuristic Analysis: Ideal Inference of DragOSM}
\label{ideal_dragosm}
This part will re-model the training and inference of DragOSM. 
To simplify the training and inference discussion of DragOSM, we simplify the movement of a whole polygon as the movement of a representative point on the polygon.  
Thus, for a given point $\mathbf{x}\in \mathbb{R}^2$ on the footprint polygon, we add a Gaussian noise $\boldsymbol{\epsilon} \sim \mathcal{N}(\mathbf{0}, \sigma^2\boldsymbol{I}_2)$ on the $\mathbf{x}$:
\begin{equation}
    \tilde{\mathbf{x}} = \mathbf{x}-\boldsymbol{\epsilon},
\end{equation}
where $\tilde{\mathbf{x}}$ is the noise sample point on the simulated OSM polygon. 

From Eq.\ref{eq:o2f}, we know that DragOSM actually is learning the noise distribution of OSM polygons, and here is an assumption based on the noising process of training: the OSM polygons are normally distributed around the ground truth footprints. To model the denoising process, we denote the DragOSM network for one noise point as $\boldsymbol{\epsilon}_\theta(\tilde{\mathbf{x}})$, which can predict an offset that has the same dimensions as $\mathbf{x}$, enabling direct correction of the input. 

In inference, for real-world OSM polygons in the test set, we assume the OSM point is also normally distributed around the footprint with distribution $\boldsymbol{\zeta} \sim \mathcal{N}(\mathbf{0}, \nu^2\boldsymbol{I}_2)$. 

\textbf{One-step Prediction.}  Under single-step inference, if an ideal DragOSM perfectly learns the Gaussian distribution relationship between OSM and footprint positions observed during training, then from the Smooth L1 Loss used in Eq.~\ref{eq:loss},
\begin{align}    
    \mathcal{L}_f(\theta)=||\tilde{\mathbf{x}} +  \boldsymbol{\epsilon}_\theta(\tilde{\mathbf{x}})-\mathbf{x}||_1=0,
\end{align}
which means, 
\begin{equation}
    \boldsymbol{\epsilon}_\theta(\tilde{\mathbf{x}}) -\boldsymbol{\epsilon}=0,
\end{equation}
and the single-step relationship for an ideal DragOSM with input $\tilde{\mathbf{y}}$ is,
\begin{equation}
    \mathbf{y}=\hat{\mathbf{y}}=\tilde{\mathbf{y}}+\lambda\boldsymbol{\epsilon}_\theta(\tilde{\mathbf{y}})
\end{equation}
where $\mathbf{y}$ denotes the ground-truth footprint position in the test set, while $\hat{\mathbf{y}}$ represents the model’s estimator for $\mathbf{y}$ given the input $\tilde{\mathbf{y}}$, and $\lambda$ is a needed length compensation bridging the gap between the training distribution and the distribution of the test set. Specifically, if $\nu,\sigma$ were fixed, then $\lambda=\frac{\nu}{\sigma}$, which makes $\boldsymbol{\zeta}=\lambda\boldsymbol{\epsilon}$.

\textbf{Multi-step Prediction.}
However, in real-world training scenarios, the aforementioned ideal assumptions may not hold. For example, while the noise added during training is sampled as $\boldsymbol{\epsilon} \sim \mathcal{N}(\mathbf{0}, \sigma^2\boldsymbol{I}_2)$, the variance learned by the model may converge to $\hat{\sigma}^2$, and $\hat{\sigma}\neq\sigma$, which may differ from the true training variance. Furthermore, the noise parameter in the test set, $\boldsymbol{\zeta} \sim \mathcal{N}(\mathbf{0}, \nu^2\boldsymbol{I}_2)$, is generally unknown. 
Therefore, let us consider a sequence of positions $\tilde{\mathbf{y}}_0, \ldots, \tilde{\mathbf{y}}_t, \ldots, \tilde{\mathbf{y}}_T$, where $\tilde{\mathbf{y}}_0$ denotes the initial position $\tilde{\mathbf{y}}$. Under the framework of Eq.\ref{eq:denoise_step}, we associate this with a sequence of step sizes $a_1, \ldots, a_t, \ldots, a_T$, and the relationship are as follow:
\begin{equation}
\label{eq:single_step_point}
    \tilde{\mathbf{y}_t} = \tilde{\mathbf{y}}_{t-1}+a_t\cdot\boldsymbol{\epsilon}_\theta(\tilde{\mathbf{y}}_{t-1}),
\end{equation}
where $a_t$ is the defined step length in Eq.\ref{eq:denoise_step}.
Essentially, Eq.\ref{eq:single_step_point} assumes that, for the test set, the displacement between the annotated and true footprint positions follows a Gaussian distribution. Under this assumption, $\boldsymbol{\epsilon}_\theta(\tilde{\mathbf{y}}_{t-1})$ in Eq.\ref{eq:single_step_point} can be interpreted as the derivative of the Gaussian process describing the transformation from the footprint location to the OSM annotation. 

According to statistical theory, the derivative of a Gaussian process is a Gaussian process, provided that the covariance function is twice differentiable. Moreover, any linear combination of Gaussian processes also yields a Gaussian process. 
Based on our design, the misplaced polygons in training are generated by an additive Gaussian noise and the ground truth locations. 
The task for the model is to learn a continuous Gaussian mapping from the current position to the ground-truth center; however, the results of this training are not ideal in practice.
This implies that we can only model the model's denoising function, $\boldsymbol{\epsilon}_\theta(\tilde{\mathbf{y}}_{t})$, as being proportional to the true noise $\boldsymbol{\epsilon}$, rather than perfectly equal to it, \ie, $\boldsymbol{\epsilon}_\theta(\tilde{\mathbf{y}}_{t}) \propto \boldsymbol{\epsilon}$.

Under this idealized assumption, the offset predicted by the model, $\boldsymbol{\epsilon}_\theta(\tilde{\mathbf{y}}_{t})$, follows $\boldsymbol{\epsilon}_\theta(\tilde{\mathbf{y}}_{t}) \sim \mathcal{N}(\mathbf{0}, \nu_{t}^2\mathbf{I}_2)$, where $\nu_{t}^2=b_{t}^2\nu^2$. To characterize the convergence behavior of DragOSM, we introduce a sequence $b_0, \ldots, b_t, \ldots, b_T$ to reflect the convergence of DragOSM, where $b_t\geq0$. In particular, for any sequence $\{a_t\}_{t=1}^T$ that ensures model convergence, we have $b_t < 1$.

In other words, for Eq.\ref{eq:single_step_point} and Eq.\ref{eq:o2f_test}, we have:
\begin{equation}
\label{eq:single_step_point}
    \hat{\mathbf{y}}=\tilde{\mathbf{y}_t} - \tilde{\mathbf{y}}_{0}=\sum_{t=1}^Ta_t\cdot\boldsymbol{\epsilon}_\theta(\tilde{\mathbf{y}}_{t-1}).
\end{equation}
Based on the above assumption and the additivity of Gaussian distributions, the distribution of estimated $\hat{\mathbf{y}} \sim \mathcal{N}(\mathbf{0}, \sum_{t=1}^Ta_t^2b^2_{t-1}\nu^2\boldsymbol{I}_2)$. 
This implies that, when DragOSM reaches convergence in multi-step inference, there exists an implicit relationship among the model’s intrinsic convergence, the step size compensation, and the assumed distribution of the true test set:
\begin{equation}
\label{eq:ideal_multi_step_inference}
    \nu^2\sum_{t=1}^Ta_t^2b^2_{t} \rightarrow \nu^2\Leftrightarrow f(T)=\sum_{t=1}^Ta_t^2b^2_{t} \rightarrow 1,
\end{equation}
where we define the sequences $a_t^2 = A[t]$ and $b_t^2 = B[t]$. The value of $B[t]$ depends on the relative position between input $\tilde{y}_{t-1}$ and the mean center, and ideally approaches zero as it gets closer to the ground truth footprint position. Therefore, the convergence of the model’s denoising inference is equivalent to the convergence of $f(T)$, which in turn reduces to comparing the convergence properties of the series $\sum^T_{t=1}A[t]$ and $\sum^T_{t=1}B[t]$.

For the relationship in Eq.\ref{eq:ideal_multi_step_inference}, there are two conclusions. First, an ideal DragOSM always has $f(T)\rightarrow1$, when DragOSM is under settings that can perfectly drag the OSM to the related footprints.

\begin{figure}
    \centering
    \includegraphics[width=1\linewidth]{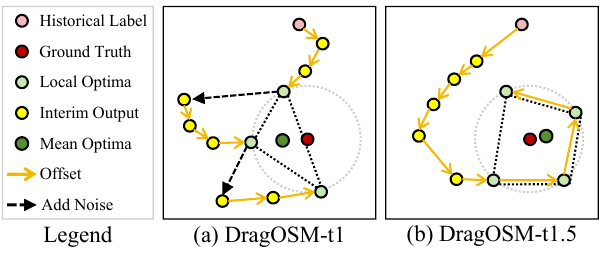}
    \caption{An illustration of proposed DragOSM-t1 and DragOSM-t1.5 in inference stage. }
    \label{fig:t1}
\end{figure}

Second, let $A[t]=c$, where $c$ is a constant, and consider a given $t=n$. At this point, the input position $\mathbf{y}_n$ reaches a local optimum under the setting of $A[t]=c$. Since $B[t]$ depends only on the distance from the input $\mathbf{y}_t$ to the center (the $\mathbf{0}$ center of the Gaussian distribution), we have:
\begin{equation}
\label{eq:local_optim}
    \lim_{T\rightarrow\infty} \frac{1}{T-n+1}f(T)=\lim_{T\rightarrow\infty} \frac{c\sum^T_{t=n}B[t]}{T-n+1}\rightarrow C,
\end{equation}
where $C$ is another constant. 
Eq.\ref{eq:local_optim} implies that, in the physical space, as $T$ increases, the position updates of $y_T$ will no longer approach the mean center, but rather oscillate along a circular path centered at the mean center. 
The convergence of Eq.\ref{eq:local_optim} can also be observed in Fig.\ref{fig:scatter}. 
Based on this observation, in Sec.\ref{tta}, we attempt to improve model performance by extending the inference time and leveraging the spatial dynamics of $\mathbf{y}_T$ during its updates.

\subsection{Test-Time Augmentation}
\label{tta}

From Fig. \ref{fig:DragOSM_inference}, DragOSM is a label-in-label-out model. Therefore, inspired by the Chain of Thought (CoT) paradigm in Multimodal Large Language Models~\cite{testscaling}, we aim to achieve improved results by synthesizing the outputs from multiple rounds of inference.

As illustrated in Fig. \ref{fig:t1}, we designed two advanced inference strategies to obtain a more robust final output by aggregating multiple candidate solutions (``local optimal points''). For the experiments reported in Table~\ref{tab:footprint_results}, we evaluate these two variants, DragOSM-t1 and DragOSM-t1.5, both of which use a 5-step inference as their base unit: (1) DragOSM-t1 employs a multi-run strategy to escape local optima. After a 5-step inference cycle, it adds random noise to the resulting polygon and uses this perturbed polygon as the input for a new 5-step cycle. This process is repeated, and the final output is the average of the positions obtained at the end of each independent run.
(2) DragOSM-t1.5 is based on the observation in Sec.\ref{dis_inference} that the model tends to converge after the initial steps, so we run the model for an additional 5 steps. The final prediction is then calculated by averaging the intermediate results from only these later steps (\ie, steps 6-10).

As the results confirm (Table~\ref{tab:footprint_results}), both the t1 and t1.5 strategies can improve the model's overall performance. Furthermore, these iterative strategies are more efficient than simply increasing the total number of inference steps.

\begin{table}
    \centering
    \caption{Performance comparison of DragOSM and t1, t1.5 versions on footprint mask metrics, MF and MI on ReBO dataset.}
    \label{tab:footprint_results}
    \begin{tabular}{llcccccc}
        \toprule
        {Model} & $T$ & {F1} & {Prec.} & {Rec.} & {IoU} & {MF} & {MI} \\
        \midrule
        \multirow{4}{*}{Base} 
            & 1  & 80.71 & 81.23 & 80.23 & 69.29   & 83.21  & 72.64 \\
            & 5  & 91.74 & 92.03 & 91.47 & 85.24   & 91.55  & 84.83 \\
            & 10 & 92.39 & 92.69 & 92.12 & 86.27   & 91.98  & 85.51 \\
            & 30 & 92.36 & 92.64 & 92.09 & 86.19   & 92.14  & 84.98 \\
        \midrule
        \multirow{4}{*}{t1} 
            & +5  & 92.39 & 92.68 & 92.13 & 86.32   & 92.12  & 85.78 \\
            & +10 & 92.49 & 92.78 & 92.23 & 86.50   & 92.27  & 86.03 \\
            & +15 & 92.58 & 92.87 & 92.30 & 86.61   & \textbf{92.35}  & \textbf{86.16} \\
            & +20 & 92.55 & 92.85 & 92.28 & 86.57   & 92.34  & 86.13 \\
        \midrule
        t1.5 & +5 & \textbf{92.67} & \textbf{92.94} & \textbf{92.42} & \textbf{86.75} & 92.25 & 85.96 \\
        \bottomrule
    \end{tabular}
\end{table}

\subsection{Extra Benefits of DragOSM}
\label{benefit}
In addition to achieving higher roof–footprint mask accuracy, the DragOSM approach offers several advantages that are not available in other methods. Most notably, it preserves the spatiotemporal continuity of label–image relationships across different periods, as shown in Fig.\ref{fig:vis_off_nadir} and Fig.\ref{fig:vis_near_nadir}. The updated labels produced by DragOSM naturally maintain this continuity, whereas extraction-based algorithms often suffer from the inherent randomness of segmentation outputs, \eg, multiple predictions for a single building in different images derived from diverse shooting angles and satellite sensors, which makes it challenging to update spatial relationships consistently. This continuity is crucial for downstream building monitoring tasks. As an example, in land survey applications such as illegal building detection, final decisions in specific cases often rely on manual verification of the correspondence between imagery and annotation.

Furthermore, the polygonal label-in, label-out nature of DragOSM not only leads to higher accuracy and better contour matching for buildings but also offers practical advantages over mask-based extraction methods, \eg, more efficient storage and greater ease of editing.
By fully leveraging existing human-labeled data, this approach enhances the value and efficiency of previous annotation work. Additionally, this work provides a novel and low-cost algorithmic approach for the maintenance of open vector maps: updating vector data through multi-round interactions between labels and imagery.

\subsection{Limitations and Future}
\label{limits_future}
Rapidly obtaining roof-footprint annotations for a given region using DragOSM represents a fundamental task, yet many aspects still need further development and improvement in this area.

Roof–footprint extraction algorithms based on roof–footprint similarity are generally founded on the common assumption that roofs and footprints share similar shapes~\cite{BONAI, zhou2025nesf, MTBRNet, obm, pang2023detecting}. The design of the DragOSM algorithm also leverages similar assumption relationships inherent in OSM annotations. 
However, a minority of buildings, due to unique architectural styles introduced by different designers, may not satisfy this assumption. Additionally, while historical annotations can provide relatively reliable building label information, the quality of some labels (\eg, OSM annotations contributed by volunteers or data authorized by government agencies) can vary significantly. Therefore, after achieving basic positional alignment, further refinement of label shapes from single or multi-view imagery represents a promising direction for future research. 

Although the ReBO dataset is a significant first step, covering 41 cities, data for this pioneering research area remains scarce. As a study based on aligning vector annotations with Very High-Resolution imagery, the availability of suitable label-image pairs is still limited. To enable broader research into the application of such alignment algorithms, the creation and contribution of more datasets by the wider research community are essential.

On the other hand, after achieving accurate alignment of building annotations, there is potential for further algorithm advances in several areas, \eg, change detection across temporal off-nadir images and labels~\cite{pang2023detecting}, as well as monocular 3D reconstruction of buildings~\cite{3d_building_reconstruct, MLS-BRN, MTBRNet}, depending on the reliability of the annotations themselves. Meanwhile, large-scale urban studies that directly utilize historical labels can, in the future, benefit from DragOSM to obtain accurately matched image–label data sources, even in cases where labeling errors may have previously been ignored for the inconvenience of correcting them. 


\section{Conclusion}

In this paper, we addressed the challenging problem of aligning historical building labels with contemporary, off-nadir remote sensing imagery. We introduced DragOSM, a novel framework that reframes this challenge as a sequential alignment task. By proposing the ``alignment token'' to explicitly model spatial discrepancies and adopting a robust noise-and-denoise training paradigm, our method effectively corrects positional errors between historical vector data and the buildings in new images.
Our extensive experiments, conducted on the newly created ReBO benchmark dataset, have validated this approach. The results demonstrate that DragOSM not only significantly outperforms existing methods but also establishes a new state-of-the-art performance baseline. The effectiveness of the alignment token and the denoising strategy confirms that modeling label correction as an iterative refinement process is a powerful and viable solution. 
\bibliographystyle{IEEEtran}
\bibliography{main.bib} 

\begin{IEEEbiography}[{\includegraphics[width=1in,height=1.25in,clip,keepaspectratio]{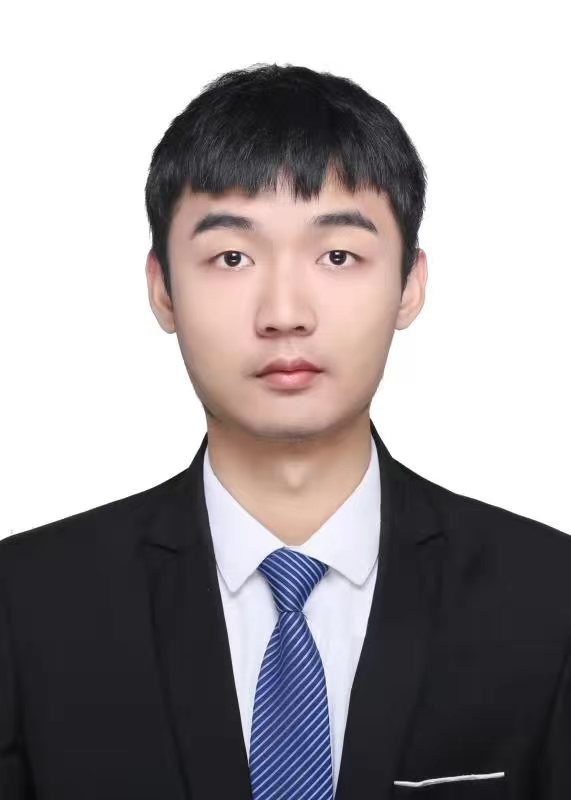}}]{Kai Li}\\
(Graduate Student Member, IEEE)
    received a B.S. degree from the University of Electronic Science and Technology of China, Chengdu, China, in 2021. 
    He is a Ph.D. candidate in Signal and Information Processing at the University of Chinese Academy of Sciences (UCAS), Beijing, China, under the supervision of Prof. \href{http://english.aircas.cas.cn/education2022/supervisors/202211/t20221104_322832.html}{Zhongming Zhao} and Prof. Yu Meng. He is also a joint Ph.D. student in Data Science at the City University of Hong Kong, advised by Prof. \href{https://zhaoxyai.github.io}{Xiangyu Zhao}. His research focuses on computer vision, especially off-nadir related topics.
\end{IEEEbiography}

\begin{IEEEbiography}[{\includegraphics[width=1in,height=1.25in,clip,keepaspectratio]{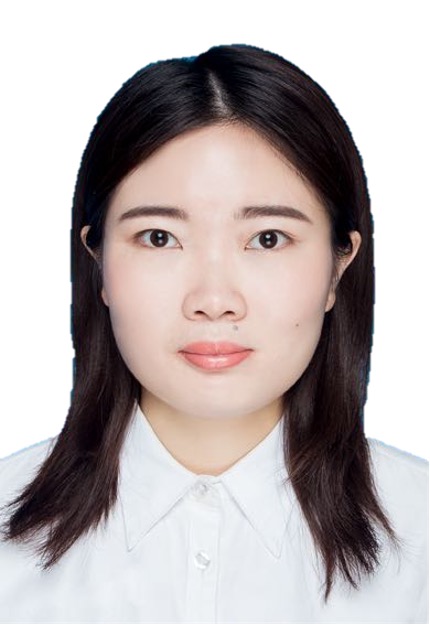}}]{Xingxing Weng} received her B.S. and M.S. degrees from the School of Remote Sensing and Information Engineering, Wuhan University, Wuhan, China, in 2017 and 2020, respectively, where she is currently pursuing a Ph.D. degree in computer science and technology with the School of Computer Science. Her research interests include remote sensing image interpretation, incremental learning, and computer vision.
\end{IEEEbiography}

\begin{IEEEbiography}[{\includegraphics[width=1in,height=1.25in,clip,keepaspectratio]{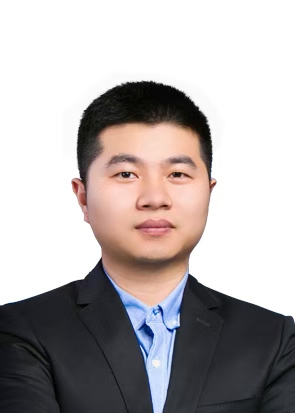}}]{Yupeng Deng} received the Ph.D. degree from the Aerospace Information Research Institute, Chinese Academy of Sciences, Beijing, China, in 2023. He is currently a special assistant Professor with the Aerospace Information Research Institute, Chinese Academy of Sciences. His research interests include computer vision, remote sensing intelligent mapping and change detection.
\end{IEEEbiography}

\begin{IEEEbiography}[{\includegraphics[width=1in,height=1.25in,clip,keepaspectratio]{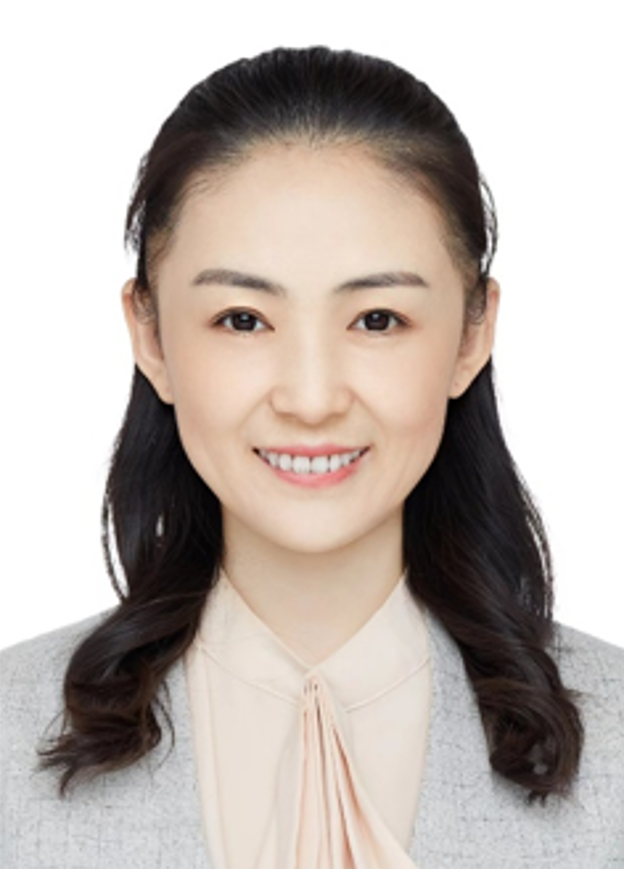}}]{Yu Meng} received the Ph.D. degree in signal and information processing from the Institute of Remote Sensing Applications, Chinese Academy of Sciences, Beijing, China, in 2008.
She is currently a professor at the Aerospace Information Research Institute, Chinese Academy of Sciences. Her research interests include intelligent interpretation of remote sensing images,remote sensing time-series signal processing, and big spatial-temporal data application.
Dr. Meng serves as an editor and board member of the National Remote Sensing Bulletin, Journal of Image and Graphics.
\end{IEEEbiography}

\begin{IEEEbiography}[{\includegraphics[width=1in,height=1.25in,clip,keepaspectratio]{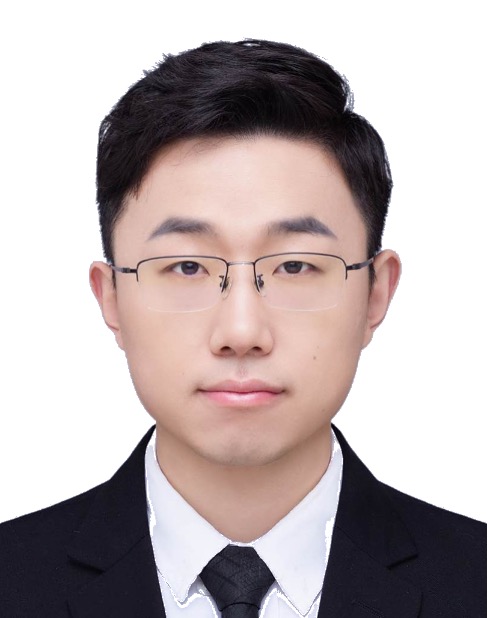}}]{Pang Chao} received the B.S. degree from China University of Mining and Technology, Xuzhou, China, in 2017, his M.S. degree from the School of Geodesy and Geomatics, Wuhan University, Wuhan, China, in 2020, and the Ph.D. degree from the School of Computer Science, Wuhan University, in 2024. He is currently a postdoctoral fellow with the School of Artificial Intelligence, Wuhan University. His research interests include time-series remote sensing image analysis and large multimodal models.
\end{IEEEbiography}

\begin{IEEEbiography}[{\includegraphics[width=1in,height=1.25in,clip,keepaspectratio]{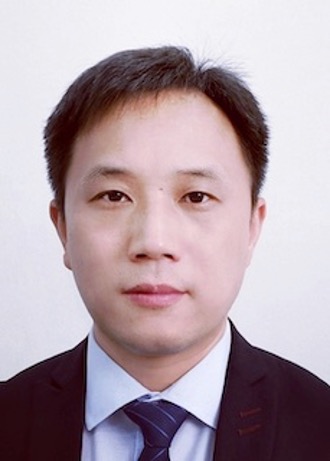}}]{Gui-Song Xia}
(Senior Member, IEEE) received the Ph.D. degree in image processing and computer vision from CNRS LTCI, T\'el\'ecom ParisTech, Paris, France, in 2011. From 2011 to 2012, he has been a postdoctoral Researcher with the Centre de Recherche en Math\'ematiques de la Decision, CNRS, ParisDauphine University, Paris, for one and a half years. He is currently working as a full professor in computer vision and photogrammetry with Wuhan University. He has also been working as a visiting scholar at DMA, \'Ecole Normale Sup\'erieure (ENS-Paris) for two months, in 2018. He is also a guest professor of the Future Lab AI4EO in Technical University of Munich (TUM). His current research interests include mathematical modeling of images and videos, structure from motion, perceptual grouping, and remote sensing image understanding. He serves on the Editorial Boards of several journals, including \textit{ISPRS Journal of Photogrammetry and Remote Sensing, Pattern Recognition, Signal Processing: Image Communications, EURASIP Journal on Image \& Video Processing, Journal of Remote Sensing, and Frontiers in Computer Science: Computer Vision.}
\end{IEEEbiography}

\begin{IEEEbiography}[{\includegraphics[width=1in,height=1.25in,clip,keepaspectratio]{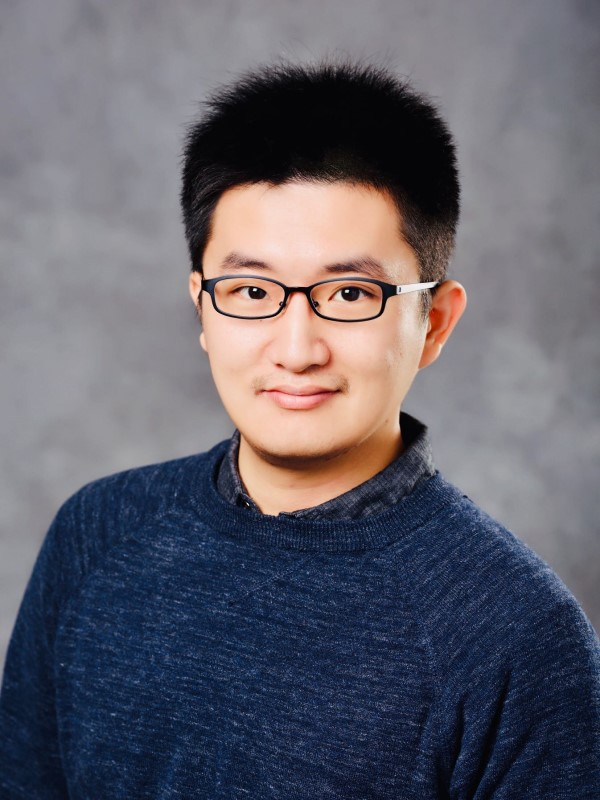}}]{Xiangyu Zhao} is an assistant professor of the school of data science at City University of Hong Kong (CityU). Prior to CityU, he completed his PhD (2021) at MSU, MS (2017) at USTC and BEng (2014) at UESTC. 
His current research interests include data mining and machine learning. He has published more than 100 papers in top conferences (\eg, KDD, WWW, AAAI, SIGIR, IJCAI, ICDE, CIKM, ICDM, WSDM, RecSys, ICLR) and journals (\eg, TOIS, SIGKDD, SIGWeb, EPL, APS). 
His research has been awarded ICDM'22 and ICDM'21 Best-ranked Papers, Global Top 25 Chinese New Stars in AI (Data Mining). He serves as top data science conference (senior) program committee members and session chairs (\eg, KDD, WWW, SIGIR, IJCAI, AAAI, ICML, ICLR), and journal guest editors and reviewers (\eg, TKDE, TKDD, TOIS, TIST, CSUR, Frontiers in Big Data). 
He serves as the organizer of Tutorial Co-Chair@APWeb-WAIM'24, DRL4KDD@KDD'19 / WWW'21, DRL4IR@SIGIR'20-22 / CIKM'23, DLP@RecSys'23, RWGM@WWW'24, and a lead tutor at KDD'23, WWW'21/22/23, IJCAI'21/23, and WSDM'23, etc. More information about him can be found at \url{https://zhaoxyai.github.io/}.
\end{IEEEbiography}

\end{document}